\documentclass{article}

\usepackage{microtype}
\usepackage{graphicx}
\usepackage{subfigure}
\usepackage{booktabs} % for professional tables
\usepackage[T1]{fontenc}
\usepackage{lipsum}
\usepackage{hyperref}

\usepackage{amsmath}
\usepackage{amssymb}
\usepackage{mathtools}
\usepackage{amsthm}

\usepackage[capitalize,noabbrev]{cleveref}

\theoremstyle{plain}

\theoremstyle{definition}

\theoremstyle{remark}

\usepackage{url}
\usepackage[utf8]{inputenc}
\usepackage{inconsolata}
\usepackage[normalem]{ulem}

\usepackage{xspace}
\usepackage{bbm}
\newcommand*{\neuronvote}{\texttt{NeuronVote}}
\newcommand*{\avgoverlap}{\texttt{AvgOverlap}}

\usepackage[final]{neurips_2023}

\usepackage[utf8]{inputenc} % allow utf-8 input
\usepackage[T1]{fontenc}    % use 8-bit T1 fonts
\usepackage{hyperref}       % hyperlinks
\usepackage{url}            % simple URL typesetting
\usepackage{booktabs}       % professional-quality tables
\usepackage{amsfonts}       % blackboard math symbols
\usepackage{nicefrac}       % compact symbols for 1/2, etc.
\usepackage{microtype}      % microtypography
\usepackage{xcolor}         % colors

\usepackage[colorinlistoftodos,prependcaption,textsize=small]{todonotes}
\presetkeys{todonotes}{inline}{}

\title{Evaluating Neuron Interpretation Methods \newline of NLP Models}

 \author{Yimin Fan$^{\heartsuit}$ \hspace{2mm} 
 Fahim Dalvi$^{\diamondsuit}$ \hspace{2mm}
 Nadir Durrani$^{\diamondsuit}$ \hspace{2mm}  
 Hassan Sajjad$^{\clubsuit}$\\
 $^{\heartsuit}$The Chinese University of Hong Kong, Hong Kong, China \\ 
 $^{\diamondsuit}$Qatar Computing Research Institute, HBKU, Qatar \\
 $^\clubsuit$Faculty of Computer Science, Dalhousie University, Canada \\
}
\begin{document}

\maketitle

\begin{abstract}

Neuron interpretation offers valuable insights into how knowledge is structured within a deep neural network model. While a number of neuron interpretation methods have been proposed in the literature, the field lacks a comprehensive comparison among these methods. This gap hampers progress due to the absence of standardized metrics and benchmarks. The commonly used evaluation metric has limitations, and creating ground truth annotations for neurons is impractical. Addressing these challenges, we propose an evaluation framework\footnote{\url{https://github.com/fdalvi/neuron-comparative-analysis}} based on voting theory. Our hypothesis posits that neurons consistently identified by different methods carry more significant information. We rigorously assess our framework across a diverse array of neuron interpretation methods. 
Notable findings include: i) despite the theoretical differences among the methods, neuron ranking methods share over 60\% of their rankings when identifying salient neurons, ii) the neuron interpretation methods are most sensitive to the last layer representations, iii) Probeless neuron ranking emerges as the most consistent method.
\end{abstract}

%\vspace{-2mm}
\section{Introduction}
%\vspace{-1mm}

The advent of deep neural network (DNN) models and their success in Natural Language Processing (NLP) has opened 
a new research frontier to interpret the representations learned within these models. 
Representation analysis provides a holistic view of the knowledge learned within the representation. Whereas neuron analysis gives insight into how knowledge is structured within the representation. To this end, an ample amount of work has been done to understand the knowledge captured within the learned representations \citep{belinkov:2017:acl,liu-etal-2019-linguistic, tenney-etal-2019-bert,dalvi2022discovering} and individual neurons~\citep{karpathy2015visualizing, kadar-etal-2017-representation, dalvi:2019:AAAI}. 

Neuron analysis provides a fine-grained interpretation of the representation. 
%The goal of 
%Representation analysis 
%is to 
%provides a holistic view of the knowledge learned within the representation. On the contrary, neuron analysis gives insight into how knowledge is structured within the representations. 
It discovers salient neurons of the network w.r.t to a concept such as morphological or syntactic concepts~\citep{lakretz-etal-2019-emergence, durrani-etal-2020-analyzing}, and provides insight into how they may be used during inference~\citep{shappely_NIPS2017_7062,kamdhere}. Beyond understanding the inner dynamics of the DNN models, neuron analysis has various potential applications such as network manipulation~\citep{bau2018identifying}, domain adaptation \citep{IDANI}, model distillation~\citep{dalvi-2020-CCFS} and architectural search~\citep{prasanna-etal-2020-bert}. %In this work, we focus on neuron analysis methods.

A number of neuron analysis methods have been proposed in the literature to interpret deep NLP models. While these methods have shown promising results, there is a notable absence of systematic comparisons among them in existing studies. %A fair comparison across neuron interpretation methods is important to understand the strengths and weaknesses of each method and to guide the community toward developing better interpretation methodologies.
Conducting a fair comparison of neuron interpretation methods is crucial for comprehending the merits and limitations of each technique. Such comparisons serve as a guiding light for the community, steering efforts towards the development of improved and more effective interpretation methodologies.
%critical for the progress of the field, and it is 

In this work, we study a variety of neuron interpretation methods under a controlled setting and provide a thorough comparative analysis. There are two main challenges in comparing neuron interpretation methods: i) absence of a \emph{standard evaluation metric}, ii) lack of an \emph{evaluation benchmark}~\citep{neuronSurvey}. The most common evaluation metric used in the neuron analysis literature is to train a classifier using the discovered neurons as features. The performance of the classifier in predicting the concept of interest serves as evidence of the quality of the discovered neurons, in other words, the correctness of the neuron interpretation method. 

We argue that the classifier as an evaluation metric is suboptimal for two reasons: i) training a classifier provides an opportunity to learn the task and it may result in good performance irrespective of the quality of the neurons used as features~\citep{hewitt-liang-2019-designing}, ii) the classifier as an evaluation method may favor the neuron interpretation methods that are methodologically similar (for instance, methods that use a classifier to discover salient neurons). \cite{antverg2021pitfalls} shed light on the latter issue and showed that the methodological similarity between the neuron interpretation methods and the evaluation methods may result in unreasonably high accuracy scores. We empirically demonstrate the issue in Section~\ref{sec:evaluation_methodology}. Another bottleneck in comparing neuron interpretation methods is the absence of \emph{evaluation benchmark}. Neurons are distributive in nature and exhibit polysemy (capturing multiple concepts at the same time). Moreover, there can be different sets of neurons learning a
%single 
concept, and there exists no single correct answer. It is therefore intricately challenging to create the ground truth annotations 
%of neurons 
and infeasible to scale 
%the effort 
to a large set of models and concepts. 

Given the lack of a standard evaluation metric and the infeasibility of creating ground truth annotations, \emph{how can we compare neuron interpretation methods?} We rely on the voting theory~\citep{voting_theory} that performs a systematic aggregation of results in order to achieve a consensus~\citep{consensus_inbook}. It is effectively used in political, social, and machine-learning communities for cases wherein there are distinct preferences among voters (models). Examples of specific applications of voting theory include ranking search results obtained from various algorithms in information retrieval~\citep{consensus_ranking,rank_correlation_ir}, finding the best machine translation system using pair-wise ranking~\citep{lapata-2006-automatic} and aggregating results of several NLP tasks~\citep{nlp_voting_benchmarking}.

%\hs{Done. remove Wikipedia links and add citations. Talk about rank correlation vs. set based methods}

%We consider each neuron interpretation method as a voter that votes for every neuron with respect to a concept. 
%
%In this work, we target the problem of the lack of an evaluation metric and the complexity of developing a ground truth annotation. \nd{This is sudden. We need to clearly establish that bench-marking is infeasible and somehow motivate that we are presenting Voting as a guiding principal for the researchers } 
%Given a neuron interpretation method under observation, one may take the Majority Rule, a principle that the decision-making is done by the group that has the most members. 
%\emph{Voting} is a known methodology in the machine learning community where classifiers vote towards a class and majority wins. \nd{colloquial use of language} 
%
We hypothesize that \emph{neurons that are commonly discovered by different interpretation methods are more informative than others}, and may serve as a signal of correctness for the selected neurons. Motivated by this, we devise two voting-based compatibility metrics. Our first metric, \avgoverlap{}, is based on plurality voting~\citep{cumulative_voting}.
%\footnote{\url{https://en.wikipedia.org/wiki/Plurality_voting}} 
\avgoverlap{} computes an average overlap of the discovered neurons across all voters. Our second method, \texttt{NeuronVote}, inspired by Borda Count~\citep{voting_book}, takes into account the ranking of neurons of each method while calculating the average pair-wise intersection over union with all the other methods. The compatibility metrics provide a single score for each method, where a higher score refers to a neuron interpretation method whose discovered neurons are more compatible with the other methods. 
%\hs{clarfiy the following line}
%Note that our compatibility metrics are not a measure of goodness of ranking, it is an indicator of how compatible is the ranking with respect to a diverse set of possible solutions to the problem.
We further extend the evaluation of neuron interpretation methods to pair-wise comparison, providing insights into how any two given methods relate in terms of their discovered neurons. 

%In the absence of a ground truth, we leverage the neurons discovered by a diverse set of neuron interpretation methods as the silver standard.

We conduct a comparative analysis of six neuron interpretation methods using diverse concepts consisting of morphological, semantic, and syntactic properties and using three pre-trained models. 
%: compatibility metrics and pair-wise comparison. These include probing-based algorithms such as Linguistic Correlation Analysis~\citep{dalvi:2019:AAAI} and corpus-based methods such as Probeless~\citep{antverg2022on}.
%We further performed a thorough qualitative comparison of each method and provide insights into the theoretical and empirical relation between neuron interpretation methods. 
%\hs{Done: update information on other models and other tagsets}
The analysis suggests that the best performing methods, despite their methodological differences, share more than 60\% of the top neurons. Probeless ranking \cite{antverg2022on} consistently exhibited the highest overlap with other methods.
%is the most consistent method. 
Because neurons in the last layer representations are highly correlated, they are the most challenging to interpret.
%for interpretation methods.
%across 
%representations from three pre-trained models and more than 100 diverse concepts consisting of morphological, semantic, and syntactic properties. A few notable findings are: Probeless is the most consistent method across all models, concepts and nature of representations. The classification-based methods using ElasticNet and Lasso regularizations for neuron selection are the second-best methods, however,  their efficacy drops on the representations of the last layers.   
%
%Classification-based methods, mainly LCA and Lasso regularizer \cite{lakretz-etal-2019-emergence} are less effective on last-layer representations. Probeless, LCA and Lasso share their top discovered neurons. The Gaussian method \cite{torroba-hennigen-etal-2020-intrinsic} has the least overlap of discovered neurons with other methods. 
Finally, we present a case study that demonstrates the usefulness of our evaluation methodology for any new method proposed in the future (Appendix~\ref{app:case_study}). Our contributions are as follows:
\begin{itemize}
  \setlength{\itemsep}{3pt}
  \setlength{\parskip}{1pt}
  \setlength{\parsep}{1pt}
    \item We conduct the first thorough comparative analysis of a large set of neuron interpretation methods of NLP across 3 pre-trained models and using a diverse set of linguistic concepts.
    \item Our proposed evaluation methodology consisting of two compatibility metrics and a pair-wise comparison targets one of the critical limitations of the neuron interpretation studies.
    \item We provide the evaluation methodology as a framework to facilitate future studies and comparison of methods.
\end{itemize}

% 	\begin{tabular}{c|cccccccc}
% 	\toprule
% 		Words & Obama  & receives & Netanyahu  & in  & the & capital & of  & USA \\ %& reading \\
% 		\midrule
% 		Suffix & -- & s & -- & -- & -- & -- & -- & -- \\
% 		POS  & NNP & VBZ & NNP & IN & DT & NN & IN & NP \\ % & VBG \\
% 		SEM  & PER & ENS & PER & REL & DEF & REL & REL & GEO \\ %& EXS \\ 
% 		Chunk & B-NP & B-VP & B-NP & B-PP & B-NP & I-NP & B-PP & B-NP \\
% 		CCG & NP & ((S[dcl]$\backslash$NP)/PP)/NP & NP  & PP/NP  & NP/N  & N  & (NP$\backslash$NP)/NP  & NP
%         \\
%         %\midrule
%       	\bottomrule
% 	\end{tabular}
	
%     \caption{Example sentences with different word-level concepts. POS: Parts of Speech tags, SEM: Semantic tags, Chunk: Chunking tags, CCG: Combinatory Categorial Grammar tags}
% 	\label{tab:example-annotation}
% \end{table*}

\vspace{-2mm}
\section{Neuron Interpretation Methods}
\label{sec:methods}
\vspace{-1mm}

%In this section, we describe all of the neuron interpretation methods we evaluate in this work, and present them in a unified mathematical framework. One of the 

The goal of neuron interpretation methods is to rank $\mathcal{N}$ neurons with respect to some concept $\mathcal{C}$. 
%The neuron interpretation methods 
These methods can be broadly classified into two classes: corpus-based methods and probing-based methods \citep{neuronSurvey}. The former discovers the role of neurons by aggregating statistics over neuron activations and the latter trains classifiers to achieve the same. 

%\hs{FD, should we introduce S sentences}
Let $\mathcal{D}$ be a dataset of sentences (represented as a list of tokens), where each token $w$ appears in a specific context/sentence and has a contextual representation associated with it. Let $z(n, w)$ be the activation of the $n^{th}$ neuron and the $w^{th}$ token in $\mathcal{D}$. The neuron $n \in \mathcal{N}$ can be from any component of the original model, such as a specific layer. Also note that the activation value $z(n, w)$ in context-dependent, so the same $w$ can result in different $z(n, w)$ values depending on context.

A concept $\mathcal{C}$ represents a property we want to discover neurons for and $\hat{\mathcal{C}}$ represents a \emph{random} concept, i.e. containing words that are not associated with the concept $\mathcal{C}$. For example, $\mathcal{C}_{country} = $\{Rome, Paris, London, New York, \dots\} represent a concept of country names and $\hat{\mathcal{C}}$ would consist of all other words in the corpus that are not country names. More formally, $\hat{\mathcal{C}} = \mathcal{D} \setminus \mathcal{C}$. 

Let $R(n, \mathcal{C})$ represents the score of a neuron $n$ with respect to a concept $C$. Each neuron interpretation method implements $R$, and provides a ranked list of neurons with respect to a concept. In the following, we present six interpretation methods\footnote{Available in the NeuroX toolkit \cite{dalvi-etal-2023-neurox}.} studied in this paper.

\vspace{-1mm}

\subsection{Corpus-based Methods}
\label{sec:corpus-based}
Corpus-based methods align neurons with a concept by accumulating co-occurrence patterns between neuron activations and the presence of the concept of interest. We use two corpus-based methods in this work presented as follows. 
%in this subsection. 

\vspace{-1mm}
\subsubsection{Probeless}
\label{sec:probeless}

The Probeless method \citep{antverg2021pitfalls} 
%is a corpus-based method that 
obtains neuron rankings based on an accumulative strategy. The score of a given neuron $n$ is defined as follows:

\setlength{\belowdisplayskip}{0pt} \setlength{\belowdisplayshortskip}{0pt}
\setlength{\abovedisplayskip}{0pt} \setlength{\abovedisplayshortskip}{0pt}
\vspace{-2mm}
\begin{equation}
	R(n, \mathcal{C}) = \mu (\mathcal{C}) - \mu(\hat{\mathcal{C}})
	\label{eq:probeless}
\end{equation}
\vspace{-2mm}

where $\mu(\mathcal{C})$ is the average of activations $z(n, w), w \in \mathcal{C}$. $\mu(\hat{\mathcal{C}})$ is the average of activations over the random concept $\hat{\mathcal{C}}$. Note that the ranking for each neuron $n$ is computed independently.

\vspace{-1mm}
\subsubsection{IoU Method}
\label{sec:iou}

\cite{Mu-Nips} proposed IoU to generate compositional explanations for neurons. For each token in the dataset, they create i) a binary mask of a neuron by thresholding activations above a percentile and ii) a binary mask of a concept by checking its presence in the token. They compute Intersection over Union (IoU) over the two masks:

\vspace{-2mm}
\begin{equation}
	R(n, \mathcal{C}) = \frac{
		\sum_w \mathbbm{1}(z(n, w) > \delta) \wedge \mathbbm{1}(w \in \mathcal{C})
		}{
				\sum_w \mathbbm{1}(z(n, w) > \delta) \vee \mathbbm{1}(w \in \mathcal{C})
		}
	\label{eq:iou}
\end{equation}

where $\mathbbm{1}$ is the indicator function. $z(n, w) > \delta$ is a binary mask over neuron activations based on the threshold $\delta$. $w \in \mathcal{C}$ is a binary mask, created by checking if the current word represents the concept $\mathcal{C}$ of interest. \cite{Mu-Nips} used IoU scores to generate alternative explanations for neurons on the task of image classification. Here we apply their method in the NLP domain to generate neuron ranking for any concept of interest. Like Probeless, the ranking for every neuron is computed independently of other neurons.

\vspace{-1mm}
\subsection{Probing Methods}
\label{sec:probing}

Probing methods train a classifier using neuron activations as features for the task of predicting the concept of interest~\citep{belinkov:2017:acl,hupkes2018visualisation}. 
The internals of the trained classifier (e.g. its weights) are used to rank the features (neurons). %The classifier itself is usually trained on a binary task to predict if the input representation belongs to the concept of interest $\mathcal{C}$. %\fd{I restored the old definition, because it's not always "weights", e.g. in gaussian}
\vspace{-1mm}
\subsubsection{Lasso Regularizer (L1)}
\cite{Radford} trained a linear classifier with L1 regularization and used the weights of the classifier as a proxy for the importance of neurons for the given concept. The loss function of the classifier is as follows: 

\setlength{\belowdisplayskip}{0pt} \setlength{\belowdisplayshortskip}{0pt}
\setlength{\abovedisplayskip}{0pt} \setlength{\abovedisplayshortskip}{0pt}
\vspace{-2mm}
\begin{equation}
	\mathcal{L}(\theta) = -\sum_w \log P_{\theta}(c | z_\mathcal{N}(w)) + \lambda_1 \|\theta\|_1  %\nonumber
	\label{eq:lc-lasso}
\end{equation}
\vspace{-2mm}

where, $c \in \{\mathcal{C}, \hat{\mathcal{C}}\}$ is the correct class for $w$, $z_\mathcal{N}(w)$ is the vector of all neuron activations for word $w$, i.e. [$z(0, w)$, $z(1, w)$, $...$, $z(n, w)$], and $P_{\theta}(c | z_\mathcal{N}(w))$ is the probability that word $w$ belongs to the class $c$. The learned weights $\theta$ serve as the ranking of neurons. Specifically, $R(n, \mathcal{C})$ is defined as the absolute weights for the class i.e. $|\theta_\mathcal{C}(n)|$. %$\mathcal{C}$ are used to compute $R(n, \mathcal{C})$ i.e. $|\theta_\mathcal{C}(n)|$. 

% \vspace{-2mm}
% \begin{equation}
% 	R(n, \mathcal{C}) = |\theta_\mathcal{C}(n)|
% 	\label{eq:lasso}
% \end{equation}
% \vspace{-2mm}

L1 %regularization
%, alternatively called as the Lasso regularization (Least Absolute Shrinkage and Selection Operation), 
adds the ``absolute value of magnitude'' of the coefficient as a penalty term to the $L(\theta)$, which shrinks the less important weights to zero, leading to sparse weight distribution. \cite{Radford} used this regularization to force the classifier to learn spiky weights, indicating the selection of a few specialized neurons learning a concept, while setting the majority of neurons' weight to zero. Such an assumption is useful in discovering focused neurons that learned one particular concept only.

\vspace{-1mm}
\subsubsection{Ridge Regularizer (L2)}
\cite{lakretz-etal-2019-emergence} 
%alternatively 
used L2 regularization to train the linear classifier. The loss 
%function 
is as follows:
%(also known as Ridge or weight-decay regularization). This regularization adds a squared magnitude of coefficient as penalty term to the loss function:

\setlength{\belowdisplayskip}{0pt} \setlength{\belowdisplayshortskip}{0pt}
\setlength{\abovedisplayskip}{0pt} \setlength{\abovedisplayshortskip}{0pt}
\vspace{-2mm}
\begin{equation}
	\mathcal{L}(\theta) = -\sum_w \log P_{\theta}(c | z_\mathcal{N}(w)) + \lambda_2 \|\theta\|^2_2 
	\label{eq:lc-ridge}
\end{equation}

L2 forces weights to be close to zero (but not zero). It is useful to deal with multicollinearity (neurons that are highly correlated) scenarios, through constricting the coefficient while keeping all the features. Intuitively, this encourages grouping of features, thus discovering group neurons that jointly learn a concept. The score $R(n, \mathcal{C})$ is computed in a similar fashion as the Lasso Regularizer method.

\vspace{-1mm}
\subsubsection{ElasticNet Regularizier (LCA)}

Both L1 and L2 regularizations capture properties that are desirable when selecting the most important neurons. L1 facilitates sparsity, identifying focused neurons while L2 encourages identifying groups of highly correlated features into account. \cite{dalvi:2019:AAAI} used ElasticNet regularization~\citep{Zou05regularizationand} that balances the trade-off between them.  

\setlength{\belowdisplayskip}{0pt} \setlength{\belowdisplayshortskip}{0pt}
\setlength{\abovedisplayskip}{0pt} \setlength{\abovedisplayshortskip}{0pt}
\vspace{-2mm}
\begin{equation}
	\mathcal{L}(\theta) = -\sum_w \log P_{\theta}(c | z_\mathcal{N}(w)) + \lambda_1 \|\theta\|_1 + \lambda_2 \|\theta\|^2_2 %\nonumber
	\label{eq:lc}
\end{equation}

\vspace{-1mm}

$\lambda_1$ and $\lambda_2$ are hyperparameters that are tuned to optimize the effect of L1 and L2 regularization. %\cite{dalvi:2019:AAAI} used ElasticNet regularization to discover both focused %neurons highly predictive of the concept 
%and grouped neurons that jointly learn the concept.
	
\vspace{-1mm}
\subsubsection{Gaussian Classifier}
\label{ref:gaussian}
\cite{torroba-hennigen-etal-2020-intrinsic} 
%trained a generative classifier with the 
%assumption 
assumes
that neurons exhibit a Gaussian distribution. 
They 
fit a multivariate Gaussian, say $\mathcal{P}$ over all neurons. Since multivariate Gaussian is decomposable by nature, they are able to extract individual probes for any subset of input features without additional work. Formally, they are able to extract $\mathcal{P}_\mathcal{F}$, where $\mathcal{F} \subset \mathcal{N}$. The classifier itself is trained using the Bayesian framework, specifically the maximum a posteriori estimate to compute the parameters $\theta$.
Once a multivariate Gaussian is trained, the neuron selection is performed %takes places 
in a greedy fashion:
\vspace{2mm}
\begin{gather}
    \nonumber F = (), 
	\nonumber F = F \oplus \arg \max_{n} \mathcal{P}(\mathcal{C} | F \oplus \{n\}) \hphantom{...} \textrm{ for } i \textrm{ in } 1..|\mathcal{N}| \\
    R(n, \mathcal{C}) = |\mathcal{N}| - index(F, n) 
\end{gather}

In essence, probes for individual neurons are first extracted, and the neuron probe with the highest log-likelihood is considered the most important neuron. Next, probes are extracted for pairs of neurons, with one of them being the selected neuron. The pair with the highest log-likelihood then contributes the second most important neuron to the ranking. The full ranking is compared iteratively, with each step being a greedy selection of the next best neuron to add based on the log-likelihood. This method makes two fundamental assumptions: i) neuron activations follow a Gaussian distribution and ii) neuron ranking can be done in a greedy fashion. 

\vspace{-2mm}
\section{Evaluation Methodology}
\label{sec:evaluation_methodology}
\vspace{-1mm}

\subsection{Classifier Accuracy}
\label{sec:classification_acc}

The commonly used evaluation metric i.e. classification accuracy of the selected neurons is inadequate: i) because it favors the neuron ranking methods that are based on the probing framework~\citep{antverg2021pitfalls}, ii) it is not clear whether the classifier performance is a reflection of the knowledge learned in the discovered neurons or it is due to the capacity of the classifier to learn and memorize the task~\citep{hewitt-liang-2019-designing,zhang-bowman-2018-language}. 
%. This latter issue has been raised by several researchers in the work of representation analysis~\cite{hewitt-liang-2019-designing,zhang-bowman-2018-language}.
%is showcasing the knowledge learned within the representations or its capacity to memorize. These pitfalls have been repeatedly raised in the probing literature (*cite). 
%We demonstrate these issues empirically by training a classifier using the selected neurons of each method for parts of speech concepts. Table \ref{tab:sMetric2} presents the average accuracy across all parts of speech concepts when using the top 30, 50, and 100 neurons. The classifier accuracy is higher for the probing methods than the corpus-based methods, ii) probes have capacity to memorize as \textbf{Random} selection of neurons also perform reasonably well with certain number of neurons.

We demonstrate these issues empirically, by using a linear classifier without regularization as an evaluation metric 
%and use it 
to evaluate various neuron interpretation methods. Given $s$ discovered neurons with respect to a concept $\mathcal{C}$, we train the classifier using these discovered neurons as features. The performance of the classifier serves as a measure of the correctness of $s$. %We conduct experiments using the representations of 12-layered BERT-base cased model and use parts of speech (POS) tags as concepts.
%\nd{We should find a way to make this result easier for a novice reader to grasp. There is too much information that we are throwing. Why so many N? Why Random? What is NoReg? Why do we want to use it? Which of these are classifier methods? May be make the table thinner and present absolutely minimal information required to drive a point. For example we can just show Ridge, Random and Gaussian with 30 and 100 neurons.} 
Table~\ref{tab:sMetric2} %\fd{any particular reason we are leaving out LCA/gaussian? it looks weird to not include them} 
presents the average accuracy across all parts of speech (POS) concepts and across all layers of BERT for $s=30,50,70,100$. Random refers to a random selection of neurons. %NoReg uses the identical classifier for neuron interpretation as the \texttt{evaluator}. 
The details of the dataset and the experimental setup are provided in Section~\ref{sec:evaluation}. A few notable observations are: the classifier accuracy is higher for the probing methods (Lasso and Ridge) than the corpus-based methods (Probeless and IoU), ii) the classifier has the capacity to memorize as a Random selection of neurons performs reasonably well when more than a certain number of neurons are used in the evaluation.
%after a certain number of neurons, e.g. 100, \texttt{evaluator} shows above 90\% performance even when using Random selection of neurons. Other than $N=30$, NoReg is among the top two methods. Most of the classifier-based methods (Lasso, Ridge, LCA) have relatively higher performance than other methods (Probeless, IOU). 
These results complement the issues raised earlier i.e. the classifier may result in high performance irrespective of the correctness of the neurons, and it favors the interpretation methods that are methodologically similar to itself.

\begin{table}[htbp]
  \footnotesize
  \centering
  \begin{minipage}{0.45\textwidth} % Adjust the width as needed
    \centering
    \caption{\textit{Task: POS, Model: BERT,} Accuracy scores using classifier accuracy as an evaluation metric. Bold and underline are the first and second best results.}
    \vskip 0.15in

\begin{tabular}{l|cccc}
\toprule
            %& \multicolumn{6}{c}{Classification Accuracy} \\ \midrule
Neurons      %& 10    
& 30    & 50    & 70    & 100  \\ %& Avg. \\ 
\midrule
% NoReg        
% %& 0.78  
% & 0.92  & \underline{0.95}  & \underline{0.97}  & \textbf{0.98} \\ % & 0.93 \\ \midrule
Probeless   
%& 0.86  
& 0.92  & 0.94  & 0.95  & 0.96 \\ %& 0.93 \\
%Selectivity & 0.86  & 0.92  & 0.94  & 0.95  & 0.96 & 0.93 \\
IoU         
%& 0.85  
& 0.91  & 0.93  & 0.94  & 0.95 \\ %& 0.92 \\
\midrule
% Gaussian    
% %& 0.81  
% & 0.90  & 0.92  & 0.93  & 0.94 \\ %& 0.91 \\
Lasso       
%& 0.87  
& 0.93  & \underline{0.95}  & 0.95  & 0.96 \\ % & \underline{0.94} \\
Ridge       
%& 0.87  
& \textbf{0.95}  & \textbf{0.97}  & \textbf{0.98}  & \textbf{0.98} \\ %& \textbf{0.96} \\
% LCA         
% %& 0.87  
% & 0.93  & 0.94  & 0.95  & 0.96 \\ %& \underline{0.94} \\ 
\midrule
Random       
%& 0.70  
& 0.81  & 0.86  & 0.89  & 0.92 \\ %& 0.85 \\

\bottomrule
\end{tabular}

\label{tab:sMetric2}
  \end{minipage}\hfill
  \begin{minipage}{0.45\textwidth} % Adjust the width as needed
  \footnotesize
    \centering
    \caption{\textit{Task: POS, Model: BERT,} Average compatibility scores \texttt{AvgOverlap} and \texttt{NeuronVote} when selecting the top 10, 30, 50 neurons from layers 1, 6, 12. Bold 
    %numbers 
    and 
    underline 
    %numbers show 
    are
    the first 
    %best 
    %score 
    and 
    %the 
    second best scores. 
    %respectively
    }
    \begin{tabular}{l|cc}
\toprule
          & \avgoverlap{}     & \neuronvote{}     \\ \midrule
Random    & 0.021          & 0.021          \\ \hline
Gaussian  & 0.086          & 0.169          \\
LCA       & {\underline{0.258}}    & {\underline{0.514}}    \\
Lasso     & 0.240          & 0.473          \\
Ridge     & 0.177          & 0.362          \\
Probeless & \textbf{0.269} & \textbf{0.532} \\
IoU       & 0.156          & 0.365          \\ \bottomrule
\end{tabular}
\vskip -0.15in
\label{tab:comptabilityscore_avg}
  \end{minipage}
\end{table}
% \begin{table}[]
% \footnotesize
% \centering
% \caption{\textit{Task: POS, Model: BERT,} Accuracy scores using classification as an evaluation metric. Bold and underline are the first and second best results.}

% \end{table}

%In this paper, we do away from evaluating the methods. We believe that in the absence of an evaluation benchmark and standard metric, it is impossible and unfair to make this judgement. But the lack of metric should not hinder the researchers draw comparative analyses between these methods. 
\vspace{-1mm}

\subsection{Compatibility Metrics}
Given the absence of a good evaluation metric and gold annotations, we rely on consensus-based voting theory. The consensus serves as a tool to measure the relation between rankings. In NLP, it has been widely used in information retrieval to combine search results of various algorithms~\citep{consensus_ranking,rank_correlation_ir}, in machine translation to compare different systems' output using pair-wise ranking~\citep{lapata-2006-automatic} and to aggregate results of NLP tasks~\citep{nlp_voting_benchmarking}. 

One way to compare rankings is to perform pairwise comparison using rank correlation or similarity functions such as Kendall's $\tau$~\citep{kendall1938measure} and Spearman correlation~\citep{daniel1990applied}. These methods assume conjoint rankings where all rankings contain identical elements. In the case of neuron rankings, for every neuron interpretation method, we only consider the $s$ top-ranked neurons with respect to a concept. The rationale behind this setting is to minimize the noise and randomness in rankings that may occur for neurons that are not learning the concept. 
%However, 
This results in disjoint sets and the rank correlation methods can not be directly applied in such cases
%to compare such rankings 
\citep{consensus_ranking}.  

Motivated by the voting theory, we devise two metrics to compare disjoint rankings. 
%We hypothesize that \emph{neurons that are commonly discovered by different interpretation methods are more informative than others}, and may serve as a signal of correctness for the selected neurons. Motivated by this, we devise two voting-based compatibility metrics. 
Our first metric, \avgoverlap{}, is based on plurality voting~\citep{cumulative_voting}. 
%As the name suggests, 
\avgoverlap{} computes the pair-wise average neuron overlap across methods. It is a set-based method and it does not consider the internal ranking of neurons into consideration. For example, the first top neuron and fifth top neuron in a ranking will get equal weight when compared with another ranking. Our second method, \texttt{NeuronVote}, inspired by Borda Count~\citep{voting_book}, considers the order of neurons into consideration in comparing rankings. We call our methods compatibility metrics, which when given a neuron ranking, provide its compatibility score with respect to other methods. 
%
%propose two voting-based compatibility metrics to draw a comparison across neuron interpretation methods. Intuitively, 
The metrics give a high score to a method that is most aligned with the rankings of the other neuron interpretation methods. To build a deeper understanding of how any two methods relate in terms of the resulting neurons, we also present a pair-wise comparison of the discovered neurons across methods. %Our analysis provides insight into how the output of methods is related to each other. 

%
%Given the neuron scores $R(n, \mathcal{C})$, we choose a subset of top-ranked neurons from each method. 
Formally, let $S_m={n_1, n_2, ..., n_s}$ represent the $s$ top-ranked neurons for a method $m$, where $s$ is a hyperparameter that can be adjusted to vary the number of top neurons selected for evaluation. We describe the compatibility metrics and the pair-wise comparison method as follows.

\vspace{-1mm}
\subsubsection{Average Overlap}
\avgoverlap{} scores an interpretation method based on its average neuron overlap with other methods. We define the overlap $o(m_1, m_2)$ between two methods as the intersection over union of their respective top-neuron sets:

\vspace{-3mm}
\begin{equation}
	o(m_1, m_2) = \frac{|S_{m_1} \wedge S_{m_2}|}{|S_{m_1} \vee S_{m_2}|}
\end{equation}

\avgoverlap{} is defined as: % follows:
\vspace{-3mm}
\begin{equation}
	\avgoverlap_{m_i} = \frac{1}{\mathcal{M}-1}\sum_{j=1, j\neq i}^{\mathcal{M}} o(m_i, m_j)
\end{equation}

where $\mathcal{M}$ is the set of all methods.
A large value of \avgoverlap{} means that the ranking of that method is well aligned with the ranking of other neuron interpretation methods. 
% implying that it can combine the advantages of different methods.

\vspace{-1mm}
\subsubsection{NeuronVote}
\avgoverlap{} treats each neuron equally in a set of discovered neurons and ignores the ranking assigned by the 
%neuron interpretation 
methods. However, intuitively a neuron interpretation method whose top choice is endorsed by all other methods should get a higher score than an interpretation method whose 10th top choice is endorsed by all other methods. To take this into account, we follow the Borda count strategy~\citep{voting_book} and propose 
%another metric named 
\neuronvote{}. It considers each neuron's ranking in the voting process.
%
% \neuronvote{} is inspired by the concept of model ensembling in machine learning. In machine learning, aggregating the results of multiple classifiers is known to result in a stronger classifier. Similarly, neurons highlighted by multiple methods should be more informative and relevant than others. We propose a voting-based method to find these neurons. 
%Given the neurons ranked by $\mathcal{M}$ different methods as $S_{m_1}$, $S_{m_2}$ etc, each neuron is assigned votes from each method based on their ranking. More concretely, in the case of 
%
 % The first step in NeuronVote is to create a "S_best" ranking, which is derived by aggregating the weights from each neuron method. For instance, if neuron 1 has weight 10 in method 1, and weight 5 in method 2, its aggregate weight will be 15 in S_best. The intersection over union is then applied between a method's neurons and S_best. We will improve the description of the method in the final version of the paper, and formulate them mathematically for clarity. 
%
%Here, 
We first create $S_{best}$, an aggregated ranking based on the position of individual neurons in the rankings of various methods. 
\setlength{\belowdisplayskip}{0pt} \setlength{\belowdisplayshortskip}{0pt}
\setlength{\abovedisplayskip}{0pt} \setlength{\abovedisplayshortskip}{0pt}
\vspace{-1mm}
\begin{equation}
    S_{best} = {argsort}\left( \sum_{m}^{\mathcal{M}} s - index(S_{m}, n) \hphantom{...} \textrm{for } n \in \mathcal{N} \right) 
\end{equation}

where $argsort$ is an ascending sort function and $index$ gives the position of neuron $n$ in the ranked list. The accumulated $S_{best}$ represents an aggregation on weighted votes, with neurons appearing at the top of various methods' ranked lists getting a higher weight than others.
%Again, consider the neuron ranking of method $m_1$ as $S_{m_1} = n_1, n_2, ..., n_s$. The top-ranked neuron $n_1$ receives the highest weight  ($=s$) and the second-best neuron receives the second highest weight $=s-1$ and so on. The votes for each neuron are accumulated across all methods, and the neuron with the highest vote is considered as the most informative neuron. Let $S_{best}$ represent this aggregated ranking based on weighted votes.
Following the notation in \avgoverlap{}, we define $\neuronvote_{m_i}=o(S_{m_i}, S_{best})$. 
% as
% \vspace{-1mm}
% \begin{equation}
% 	\neuronvote_{m_i}=o(S_{m_i}, S_{best})
% \end{equation}
% \vspace{-1mm}
%
The method with high \neuronvote{} implies that its ranking of neurons is most endorsed by other interpretation methods. 
%prefers generally informative neurons, and thus can be considered as being closest to the "best neurons".

\vspace{-1mm}
\subsection{Pairwise Comparison}
The compatibility metrics provide a holistic evaluation of a neuron interpretation method with respect to other methods. We further extend the analysis to pair-wise comparison to understand how the discovered neurons of a method related to another method. The pair-wise comparison is calculated as an intersection between the output of two methods i.e. $o(m_1, m_2) = S_{m_1} \wedge S_{m_2}$

% \begin{equation*}
% 	o(m_1, m_2) = S_{m_1} \wedge S_{m_2}
% \end{equation*}

\vspace{-2mm}
\section{Evaluation}
\vspace{-1mm}

\label{sec:evaluation}
\subsection{Settings}
% \hs{Done: update information on other tagsets}
% \hs{Done: add information on total neurons in every layer}
% \hs{add results in appendix}

We 
%analyze the neuron interpretation methods (Section~\ref{sec:methods} using 
consider
three 12-layered pre-trained models: BERT-base-cased~\citep[BERT,][]{devlin-etal-2019-bert}, RoBERTa-base-cased~\citep[RoBERTa,][]{roberta} and XLM-Roberta-base~\citep[XLMR,][]{xlmroberta}. Each layer consists of 768 neurons.
%
%We implement all the methods in this paper in the NeuroX \cite{Dalvi_Nortonsmith_Bau_Belinkov_Sajjad_Durrani_Glass_2019} toolkit. 
%
We consider concepts from three linguistic tasks: parts of speech tags~\citep[POS,][]{marcus-etal-1993-building}, semantic tags
%using the Parallel Meaning Bank data
\citep[SEM,][]{abzianidze-EtAl:2017:EACLshort} and syntactic chunking (Chunking) using CoNLL 2000 shared task dataset \citep{tjong-kim-sang-buchholz-2000-introduction}. Dataset details are provided in Appendix~\ref{app:datastats}.
%We use the English Penn TreeBank \cite{marcus-etal-1993-building} dataset to conduct experiments. 
We consider every tag as a concept and identify neurons with respect to the concept. In the case of classifier-based methods, we use a binary classification setup where the contextualized representation of words belonging to the concept serves as positive examples. We randomly select the negative class words from the rest of the data, equal in size to the positive class examples. We drop the concepts that have less than 200 examples to ensure stable training. We split the binary classification dataset into train/dev/test splits of 70/15/15 percent. We use $\lambda_1=0.01$ and $\lambda_2=0.01$ for regularization-based methods. We present the results of POS only in the main paper and provide the rest in Appx.~\ref{app:results}.

\textbf{Number of top neurons:} We did not optimize the number of neurons $s$ selected by each method as this would result in a varying number of neurons and will make the compatibility score incompatible across methods. We consider $s$ = $10, 30, 50$ which covers a diverse range to generalize the findings. %Moreover, we consider a baseline of randomly selecting the 10, 30, and 50 neurons and calculate the compatibility score.

\textbf{Baseline:}
We generate a random ranking of neurons to compare the behavior of a neuron interpretation method with a random selection of neurons. We refer to it as \emph{Random}. %later on. 

\textbf{Procedure:}
We extract layer-wise contextualized representations of words in the dataset. For every layer representation, we discover the top neurons with respect to a concept using each neuron interpretation method. 
%Given a concept, we discover the top neurons using each neuron interpretation method. 
We follow a leave-one-out strategy %\fd{this is true for neuronVote as well? The best ranking does not include "self"?} 
to calculate the compatibility score of each method with respect to other methods. More specifically, we consider one neuron interpretation method as a test method and select the rest 
%of the methods 
as the database of the sets of top neurons. We then calculate the \texttt{AvgOverlap} and \texttt{NeuronVote} scores of the test method.

\vspace{-2mm}
\subsection{Compatibility Scores}
\vspace{-1mm}

%We follow a leave-one-out strategy to compute compatibility score of each method where other methods serve as the silver standard of comparison. For instance, in order to calculate the compatibility scores of Probeless, we consider the sets of neurons of all the other methods as the silver standard. 
%Table \ref{tab:sMetric} shows accuracy (percentage) across different methods when selecting 10--50 top neurons. These results are averaged across the understudied tags. Firstly we notice that classifier accuracy is always high for the methods that use linear probes to obtain the most salient neurons. Similarly Selectivity method gives higher score to the corpus-based method (Probeless and IoU). Notice that random neurons in the case of classifier as an evaluation metric also resulted in an accuracy greater than 85\% when more than 50 neurons are selected.  These results empirically verify the argument made by \cite{antverg2021pitfalls} who conjectured that ranking method should be disentangled from the evaluation. 
%
%
%We have established that existing metrics to measure the efficacy of the ranking are inadequate and misleading in favoring one method over another. Let us now analyze how do these methods align with each other with respect to our proposed compatibility metrics; \texttt{AvgOverlap} and \texttt{NeuronVote}. 
Table~\ref{tab:comptabilityscore_avg} presents the average compatibility scores across all concepts and when selecting the top 10, 30, and 50 neurons from layers 1, 6, and 12.

% \begin{table}[]
% \centering
% \footnotesize
% \caption{\textit{Task: POS, Model: BERT,} Average compatibility scores \texttt{AvgOverlap} and \texttt{NeuronVote} when selecting the top 10, 30, 50 neurons from layers 1, 6, 12. Bold numbers and underline numbers show the first best score and the second best score respectively}

% \vskip -0.05in
% \begin{tabular}{l|cc}
% \hline
%           & \avgoverlap{}     & \neuronvote{}     \\ \hline
% Random    & 0.021          & 0.021          \\ \hline
% Gaussian  & 0.086          & 0.169          \\
% LCA       & {\underline{0.258}}    & {\underline{0.514}}    \\
% Lasso     & 0.240          & 0.473          \\
% Ridge     & 0.177          & 0.362          \\
% Probeless & \textbf{0.269} & \textbf{0.532} \\
% IoU       & 0.156          & 0.365          \\ \hline
% \end{tabular}
% \vskip -0.15in
% \label{tab:comptabilityscore_avg}
% \end{table}

\begin{table*}[]
\centering
\footnotesize
\caption{\textit{Task: POS,} Average \neuronvote{} compatibility scores across concepts when selecting the top 10, 30, and 50 neurons from layers 1, 6 and 12. Bold numbers, underline numbers, and dashed numbers show the first, second, and third best scores respectively}
\vskip 0.05in

\begin{tabular}{l|ccc|ccc|ccc}
\toprule
          & \multicolumn{3}{c|}{BERT}                        & \multicolumn{3}{c|}{RoBERTa}                     & \multicolumn{3}{c}{XLMR}                         \\
Layers    & 1              & 6              & 12             & 1              & 6              & 12             & 1              & 6              & 12             \\ \midrule
Random    & 0.019          & 0.021          & 0.023          & 0.020          & 0.019          & 0.017          & 0.020          & 0.017          & 0.023          \\ \hline
Gaussian  & 0.147          & 0.177          & 0.183          & 0.140          & 0.179          & 0.179          & 0.118          & 0.135          & 0.153          \\
LCA       & \textbf{0.501} & \dashuline{0.544}          & \dashuline{0.391}          & \underline{0.550}    & \underline{0.495}    & \dashuline{0.380}          & \underline{0.450}    & \underline{0.531}    & \dashuline{0.373}          \\
Lasso     & \dashuline{0.496}          & \underline{0.545}    & \underline{0.395}    & \dashuline{0.475}          & 0.453          & 0.350          & \dashuline{0.410}          & \dashuline{0.471}          & 0.267          \\
Ridge     & 0.332          & 0.405          & 0.360          & 0.452          & \dashuline{0.491}          & \underline{0.510}    & 0.358          & 0.446          & \underline{0.455}    \\
Probeless & \underline{0.497}    & \textbf{0.550} & \textbf{0.515} & \textbf{0.590} & \textbf{0.589} & \textbf{0.580} & \textbf{0.534} & \textbf{0.590} & \textbf{0.497} \\
IoU       & 0.343          & 0.380           & 0.348          & 0.315          & 0.338          & 0.275          & 0.334          & 0.320          & 0.344          \\ \bottomrule
\end{tabular}

\label{tab:model_comparison}
\end{table*}

\textbf{\avgoverlap{} and \neuronvote{} show a similar pattern:} %Table~\ref{tab:comptabilityscore_avg} compares the findings of both compatibility metrics. 
While the scale of the compatibility score is different between both metrics, they are consistent in highlighting the top interpretation methods i.e. Lasso, LCA, and Probeless. In the rest of the paper, we mainly present the results of \neuronvote, and discuss the \avgoverlap{} only when it is necessary to show a different trend. 

\textbf{Overall comparison across methods:}
Table~\ref{tab:model_comparison} presents the \neuronvote{} scores averaged over all POS concepts and when selecting the top 10, 30, and 50 neurons. The bold, underline, and dashed numbers represent the first, second, and third best compatibility scores respectively. \textbf{Probeless consistently achieves the highest compatibility across all layers and models.} In other words, the neurons selected by Probeless have the highest overlap with the neurons selected by other methods. We did not observe a similar trend for IoU which is also a 
%nother 
corpus-based method. 
%studied in this work. 

\textbf{LCA and Lasso achieve
the second-best compatibility score} (see underlined and dashed numbers in Table~\ref{tab:model_comparison}). Ridge, although methodologically similar to LCA and Lasso, did not consistently show %consistently 
high compatibility %scores 
across models. We discuss these methods later in Section~\ref{sec:neuroncomparison}.  
%One corpus-based method, Probeless, and two classification-based methods, LCA and Lasso, are among the top compatible methods with respect to other neuron interpretation methods under study here (Table~\ref{tab:model_comparison} shows the top two scores, bold and underline respectively -- variation of 1 point is considered as an identical score). In other words, the neurons selected by these methods overlap the most with the sets of salient neurons of other methods. One notable point here is that despite the methodological differences in techniques, they discover a large number of identical neurons. We further discussed the overlap between each method in Section~\ref{sec:neuroncomparison}. 
%
%Ridge and IoU did not result in the top compatibility score but compared to Random, they achieved substantially high score. 
Gaussian achieved the worst compatibility score which is lower by 0.354 points from the best \neuronvote{} score on layer 1 and is only 0.128 better than Random for BERT. The result of Gaussian presents an interesting scenario where the compatibility score suggests that the neurons identified by Gaussian are least similar to all other methods. However, the difference with Random, although small, suggests that the selection of neurons is not random. The low results of Gaussian are inline with \cite{antverg2021pitfalls} where they
%compared Probeless, LCA and Gaussian, and 
found that the Gaussian method memorizes the probing task and it may not provide the most faithful ranking of neurons. These scores, therefore, suggest that further investigation may be required to confirm the efficacy of the Gaussian method. 

A notable point about the top 3 methods (Probeless, LCA, and Lasso) is their methodological diversity, despite of which they selected similar neurons to other methods. On the other hand, methodologically similar techniques such as IoU and Probeless did not result in similar compatibility scores. This shows that the compatibility metric is not biased by methodological similarity, which was one of the primary concerns with the commonly used evaluation metrics.

%Table~\ref{tab:compatibility} presents the average score across all layers and all concepts.\footnote{We experimented using different number of neurons in the range of 10 to 100 and found the results to be consistent.}. %The choice of selecting top 30 neurons is based on the 
%As we can see from the classification accuracy in Table \ref{tab:sMetric} results that top 30 neurons cover bulk ($>98\%$)of the prediction accuracy. They posses most information of the task. 

%\paragraph{Probeless, Lasso and LCA}
%Here, we discuss the results of the top three methods based on their compatibility scores. Table~\ref{tab:comptabilityscore_avg} shows the top two scores, bold and underline respectively (variation of 1 point is considered as an identical score). 
%Probeless achieved the best \avgoverlap score of 0.25 and 0.227 for layer 1 and 12, and the best \neuronvote score of 0.55 and 0.515 for layer 6 and 12. LCA achieved the best score for the remaining cases and Lasso consistently came at the second top spot (see underline values in Table~\ref{tab:comptabilityscore_avg}). 

\paragraph{Layer-wise trend:}
\cite{ethayarajh-2019-contextual} showed that each layer is different in terms of representation geometry. \cite{sajjad-etal-2022-coling-postprocessing} revealed that due to the geometry of the representation space, particularly in the last layers, knowledge of a concept may not be readily available. We hypothesize that the performance of certain neuron interpretation methods may vary based on the nature of the representation space. Our compatibility methods facilitate quantitative evidence to support this hypothesis by showing that certain neuron analysis methods suffer in selecting the most compatible neurons from the higher layers. LCA and Lasso both showed a substantial drop in their compatibility scores for the 12th layer (for example, Table~\ref{tab:model_comparison}: \neuronvote{} LCA -- 0.544 for layer 6 and 0.391 for layer 12 for BERT). In contrast, Probeless did not show any substantial difference in the scores of the last layers and earlier layers (\neuronvote: Probeless -- 0.550 vs 0.515 for layer 6 and %layer 
12 respectively for BERT). This further support Probeless as the most reliable method and encourages further investigation into the robustness of Lasso and LCA with varying representational space.
%\cite{sajjad-etal-2022-coling-postprocessing} suggested to normalize representations of higher layers for effective neuron interpretation. We did not look into this further since this is out-of-the-scope of this work.

%\hs{i forgot why we removed noReg. if we stay with this, we should remove it from the heatmap as well}

% \begin{table}[]
% \centering
% \begin{tabular}{l|ccc}
% \toprule
%           %& \multicolumn{3}{c}{Score 2 (10 neurons)} \\
% Layers    & 1            & 6            & 12          \\
% \midrule
% Random    & 0.008        & 0.016        & 0.008         \\
% \midrule
% Gaussian  & 0.176        & 0.266        & 0.214    \\
% LCA       & \textbf{0.428}        & \underline{0.456}        & 0.404      \\
% Lasso     & \underline{0.397}        & \textbf{0.459}        & \underline{0.411}  \\
% Ridge     & 0.297        & 0.315        & 0.331     \\
% Probeless & \dashuline{0.358}        & \dashuline{0.387}        & \textbf{0.459}     \\
% IoU       & 0.337        & 0.407        & 0.335       \\
% \bottomrule
% \end{tabular}
% \caption{Average \texttt{NeuronVote} score using top 10 neurons selected using the BERT model. Bold numbers and underline numbers show the first best score and the second best score respectively}
% \label{tab:comptabilityscore_10}
% \end{table}

%\nd{then we should not bring the above. It's creating an unnecessary diversion.}
%\hs{I commented out the last sentence}

\paragraph{Compatibility across models:}
The overall trend of top-performing neuron interpretation methods is similar across BERT, RoBERTa, and XLMR. One notable difference is the compatibility score of Ridge for the last layer of the RoBERTa (0.510) and XLM-R (0.439) models which is substantially higher compared to the BERT model (0.360). Moreover, it achieved the second best score for XLM-R after Probeless and is substantially better than LCA and Lasso. Ridge prefers correlated features in contrast to Lasso, which prefers a few spiky features highly predictive of the concept. Based on the results, we hypothesize that the last layers of RoBERTa and XLM-R consist of highly correlated neurons and Ridge is effective in discovering salient neurons of correlated nature. LCA balances between spiky and correlated neurons, and results in higher 
%compatibility 
scores than Lasso in these cases.

\begin{figure*}[htb]
\centering
\subfigure[Layer 1]{
\begin{minipage}[t]{0.32\linewidth}
\includegraphics[width=1\textwidth]{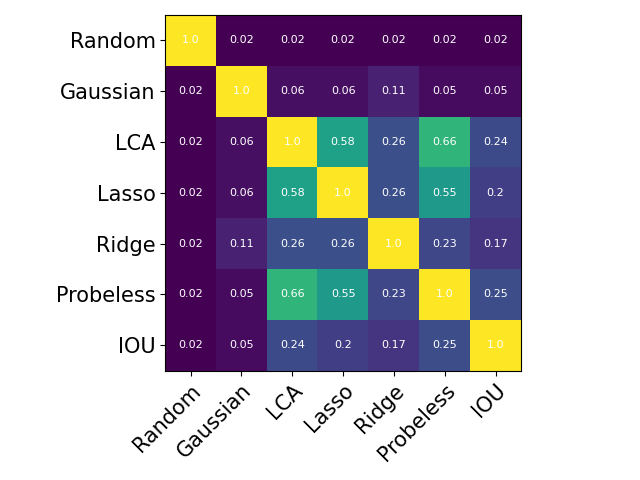}
%\caption{DT Layer 0}
\label{fig:avgl1}
\vspace{-2mm}
\end{minipage}%
}%
\subfigure[Layer 6]{
\begin{minipage}[t]{0.32\linewidth}
\includegraphics[width=1\textwidth]{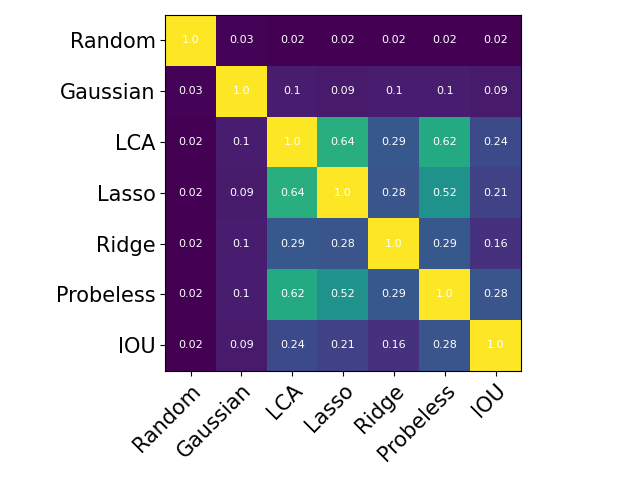}
%\caption{DT Layer 0}
\label{fig:avgl6}
\vspace{-2mm}
\end{minipage}%
}%
\subfigure[Layer 12]{
\begin{minipage}[t]{0.32\linewidth}
\includegraphics[width=1\textwidth]{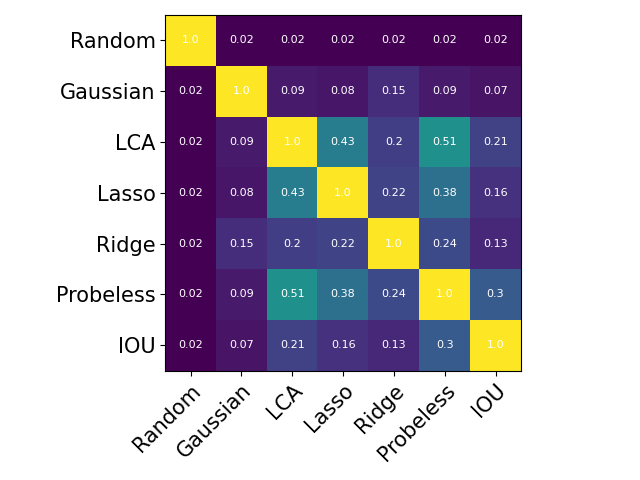}
%\caption{DT Layer 0}
\label{fig:avgl12}
\vspace{-2mm}
\end{minipage}%
}%
\vspace{-2mm}
\centering
\vspace{-2mm}
\caption{\textit{Task: POS, Model: BERT,} Comparing average overlap of top 10-50 neurons}
\label{fig:top10_50_heatmap}
\vspace{-4mm}
\end{figure*}

% \paragraph{Comparing top 10 neurons}
% We further gauge the comparison across interpretation methods when selecting the most salient 10 neurons of a concept. 
% Table~\ref{tab:comptabilityscore_10} presents the results averaged across all concepts for the BERT model. LCA showed the best compatibility for layer 1 with a margin of 0.3 points and 0.6 points over Lasso and Probeless respectively and is the most effective method in selecting the salient neurons that are  agreed by the other methods. A similar observation can be made for layer 6 where although Lasso has improved to take the first position but LCA is lowered by 0.003 points only. 
% Probeless showed substantially lower compatibility scores for layer 1 and layer 6. At layer 6, it dropped below IOU. However, at the last layer, similar to the layer-wise trend, the compatibility score of Probeless is the highest, and LCA and Lasso dropped substantially. 

\paragraph{Random selection of neurons:}
The compatibility score of Random serves as a baseline that provides a perspective of the correctness of a newly proposed neuron interpretation method. %Random resulted in 
It gave substantially lower scores than all neuron interpretation methods (Table~\ref{tab:model_comparison}). Even though the Gaussian showed the lowest compatibility score among interpretation methods, the difference with Random adds a factor of confidence to its ranking. Moreover, a low compatibility score for Random also implies that each of the other methods are bringing in important neurons to some degree. 

\vspace{-1mm}
\subsection{Pairwise Comparison}
\vspace{-1mm}
\label{sec:neuroncomparison}
The compatibility metrics provide a single score to understand an interpretation method in the context of a large set of other interpretation methods. We further extend the analysis to pair-wise comparison and provide insights into the relationship between different methods.  
Figure~\ref{fig:top10_50_heatmap} presents the heatmaps comparing the average overlap of the top 10 -- 50 neurons across methods using BERT. 
%
%\textcolor{red}{In the interest of space, we limit the results to three concepts of varying nature; DT (determiners), CD (numbers) and NNS (plural nouns), and to three layers; the first, middle and last contextualized layers.}\footnote{Appendix XX provides heatmaps of other concepts and layers with varying number of top neurons.}
%
The light color in the heatmap shows higher matches while the dark color shows lower matches. We added 
%two baselines; 
\textit{Random} -- a random selection of neurons as a baseline. %and \textit{NoReg} -- a classification-based selection of neurons without using any regularization. 
It 
%The Random neurons 
showed an average overlap of less than 3\% only 
%when compared with neuron interpretation methods 
(see the first row in heatmaps). 

\paragraph{Comparing Regularization-based Methods:}
%The substantially low overlap of the NoReg classifier with other classifier-based methods (LCA, Lasso, Ridge) depicts the importance of regularization in discovering neurons of a concept. Comparing the regularization-based methods, 
LCA and Lasso showed a high overlap of up to 64\% in the discovered neurons (see Figure \ref{fig:avgl1}, \ref{fig:avgl6}). 
%with at most 64\% overlap. %Note that L1 favors individual neurons most predictive of the concept of interest. 
However, Ridge results in a lower overlap of at most 29\% compared to other methods. Note that the Ridge regularization does not force the coefficients of neurons to zero, thus considers most of the neurons during the training phase. The resulting top neurons may represent correlated neurons capturing similar and redundant knowledge~\citep{dalvi-2020-CCFS} and may not be a representative of the neurons with the most predictive power with respect to the understudied concept. Lasso prefers individual neurons most predictive of the concept and ElasticNet (LCA) provides a good balance between selecting individual neurons and correlated neurons highly predictive of the concept. 
%We discuss these methods in relation to the other class of methods in the following paragraph.

\paragraph{Corpus-based vs. Regularization-based Methods:}
Probeless and IoU both directly rely on the activation value of a neuron. However, we did not observe a consistent overlap between their discovered neurons (at most 30\% in Figure\ref{fig:avgl12}). In comparison to other classes of methods, Probeless showed a better and consistent overlap with LCA and Lasso. For example, in Figure~\ref{fig:avgl1} we observed an overlap of 66\% between Probeless and LCA, and 55\% between Probeless and Lasso. The relatively high overlap of Probeless with methods using Lasso points towards the selection of identical focused neurons by these methods despite their methodological differences. 

% with respect to the concept.
%Probeless considers individual neurons with respect to a concept, thus it selects neurons that are individually representing a concept. The regularization-based methods consider all neurons together. The high L1 regularization results in zero coefficients for less predictive neurons (in the case of Lasso) or close to zero (in the case of LCA) and ends up selecting individual neurons that are highly predictive of the concept. LCA additionally normalizes the spiky neurons to reduce the effect of a few highly predictive neurons on the classification performance.  

%\nd{It is important to drive the discussion with take-homes, put emphasis on some results and explain what they entail. Right now its difficult to understand what value all this analysis is bringing. }

\paragraph{Outliers:}
Both Ridge and IoU show a smaller but consistent overlap with most of the methods. As discussed earlier, Ridge works well for correlated features which may not be highly predictive of the concept. IoU relies on activation values and it 
%considers a binary mask over neuron activations to identify neurons with respect to a concept. It 
may select neurons that consistently activate higher than the threshold irrespective of the concept. Gaussian did not show substantial overlap with any methods or with any combination of methods. Further investigation is needed to evaluate its' efficacy. 

\section{Related Work}

The area of interpreting deep learning models constitutes a broad expanse of research. This section provides a synthesized overview of diverse interpretability subareas within deep learning models applied to Natural Language Processing (NLP), while also outlining the scope of our study.

%\paragraph{Attribution Methods} 
\textit{Feature importance and attribution methods} 
%endeavor to 
identifies the contribution of input features to predictions. These methodologies predominantly rely on the gradient of the output concerning the input feature and determine input feature importance by evaluating the magnitude of gradient values \citep{DBLP:journals/corr/DenilDF14,integrated_gradient,danilevsky-etal-2020-survey}.

\textit{Counterfactual Intervention} revolves around an intricate analysis of the interplay between input features and predictions. This approach involves manipulating inputs and quantifying resulting output alterations. Diverse intervention strategies, including erasing input words
%, removing multiple input words, 
and substituting words with different meanings have been used ~\citep{erasure_li,10.5555/3504035.3504222}.

\textit{Attention Weights:} Numerous investigations have been directed towards interpreting components of deep learning models at varying levels of granularity. For instance, attention weights have emerged as a viable metric to gauge the interrelation between input instances and model outputs \citep{pmlr-v48-martins16,DBLP:journals/corr/abs-1904-02679}. Along these lines, \cite{geva-etal-2021-transformer} delved into the analysis of feedforward neural network components within the transformer model, revealing their functionality as key-value memories. Additionally, \cite{voita-etal-2019-analyzing} demonstrated that pruning many attention heads has minimal impact on performance.

\textit{Mechanistic Interpretability} puts a focus on reverse engineering of network weights to comprehend their behavior~\citep{nanda2023progress}. Building upon the Distill Circuits thread, \citet{elhage2021mathematical} investigated two-layered transformer models with attention blocks, identifying attention heads contributing to in-context learning. This understanding was further extended to larger transformer-based language models by \cite{olsson2022context}. To enhance neuron interpretability, \cite{elhage2022solu} introduced a Softmax Linear unit as an activation function replacement. \cite{wang2022interpretability} attempted to bridge mechanistic interpretability findings in small networks to large ones, particularly GPT-2 small. Their approach involved iteratively tracing influential model components from predictions using causal intervention. They showcased the potential of mechanistic interpretability in understanding extensive models, while also highlighting associated challenges. Similarly, \cite{bricken2023monosemanticity} uses a sparse autoencoder to disentangle polysemantic neurons.

\textit{Representation Analysis} involves probing network representations concerning predefined concepts, to quantify the extent of knowledge captured in these representations \citep{conneau2018you,liu-etal-2019-linguistic,tenney-etal-2019-bert,durrani-etal-2019-one,arps-etal-2022-probing}. This is often realized through training diagnostic classifiers for linguistic concepts, wherein classifier accuracy serves as an indicator of concept knowledge within representations. See \cite{belinkov-glass-2019-analysis} for a comprehensive survey.

\textit{Neuron Interpretation} A more intricate form of representation analysis, termed neuron interpretation, delves into how knowledge is structured within the network~\citep{neuronSurvey}. This approach establishes connections between neurons and predefined concepts, offering insights into where and how specific concept knowledge is assimilated. Work done on neuron analysis can be broadly classified into three groups: Neuron visualization 
%is a straightforward approach to understanding the function of a neuron by 
involves manual identification of patterns across a set of sentences~\citep{li-etal-2016-visualizing,karpathy2015visualizing}.
More recently \cite{foote2023neuron}
proposed an automated approach to enhance interpretability of Large Language Models (LLMs) by extracting and visualizing individual neuron behaviors as interpretable graphs. 
%However, they are infeasible to scale and the interpretation is subject to human bias.
%in the due to the need of human-in-the-loop approach. 
Corpus-based Methods explore the role of a neuron through techniques such as ranking sentences in a corpus \citep{kadar2016representation}, generating synthetic sentences \citep{poerner-etal-2018-interpretable} that maximize its activation, or computing neuron-level statistics over a corpus \citep{Mu-Nips,suau2020finding,antverg2022on}. \cite{bills2023language,mousi2023llms} proposed the use of LLM to interpret neurons. 
%an algorithm to generate neuron explanations, simulating activations using a simulator model (an LLM), and scoring the results. 
%Similarly, \cite{mousi2023llms} used LLM to interpret neurons derived from the latent space of pre-trained Language Models.
Probing Methods identify salient neurons for a concept by training a classifier using neuron activations as features 
%towards the concept on model's representation 
\citep{Radford,lakretz-etal-2019-emergence,neuronJournal} 
%or using Random Forest \cite{valipur-2019} 
or fitting a multivariate Gaussian over all neurons and then extracting individual probes for single neurons \citep{torroba-hennigen-etal-2020-intrinsic}. 

A number of works identify neurons with respect to the output class~\citep{wang-etal-2022-finding-skill,dai-etal-2022-knowledge}. They are effective in finding neurons that play a role in the prediction. 
In this paper, we focus on the neuron interpretation methods that take a concept as input and find neurons with respect to the concept. %We considered all methods mentioned in the recent survey on neuron interpretation~\cite{neuronSurvey} in our study. 
We propose an evaluation framework to formalize the evaluation and comparison of results across methods. Moreover, we propose a novel method, MeanSelect and present a case study of using the evaluation framework.

\vspace{-2mm}
\section{Conclusion and Limitation}
\vspace{-1mm}
We provided a thorough comparative analysis of neuron interpretation methods in NLP. We overcame the challenge of the lack of a standard evaluation metric and gold annotations by proposing an evaluation strategy consisting of two compatibility metrics and a pair-wise comparison. In addition to developing a capability to evaluate a new neuron analysis method, we presented various insights into existing neuron interpretation methods. For example, due to the correlated nature of last layer representations, they are most challenging for interpretation methods. The selection of top neurons overlaps substantially irrespective of the methodological differences among techniques.
 % We found that Probeless is the most consistent method across all concepts and models. LCA and L1 regularizer are the second most effective methods but their efficacy dropped for the higher layers.
 %Gaussian is the least similar to other interpretation methods. Despite the methodological differences between Probeless, and LCA, and L1, they share the most discovered neurons. 
%We further presented a case study to demonstrate the effectiveness of our evaluation methodology in analyzing a new interpretation method. 
We made the evaluation framework available to the research community. 

\textbf{Limitations:} Our methodology relied on consensus and thus inherits the limitations of that paradigm. The consensus may mislead the evaluation under certain settings e.g. if the majority of the voters form a lobby then they can skew the results of the consensus. 
%In the case of the neuron interpretation evaluation, 
%Such a 
This scenario can happen if a set of methods always produce close to identical ranking, they will cause a decrease in the compatibility score of a new method that produces a ranking different from theirs. Another possible issue is if a new method discovers an entirely different set of neurons than the ones discovered by other methods. The compatibility score of such a method will be low. We intend to mitigate these issues by including a large number of diverse neuron interpretation methods that are theoretically different from each other. We believe that under these settings, the scenario of an entirely different set of discovered neurons is more theoretical than practical. We further present a pair-wise analysis that provides an analysis at a more granular level and bring insight into how a method compares with other methods. The pairwise comparison heatmap will highlight if a set of methods form a lobby. 

Another limitation of current neuron interpretation methods is that they do not explicitly target the discovery of neurons of diverse nature such as polysemantic, and superposition. Theoretically, only the ElasticNet regularization is capable of discovering neurons learning a singular function and multiple functions. The other methods such as Probeless are incapable of discovering multifunction neurons. None of the neuron interpretation methods explicitly identify superposition neurons. Explicit modeling of neurons of different nature in a neuron interpretation method may result in discovering novel sets of neurons. The evaluation methods including our proposed framework do not explicitly compare neurons of different types.

\bibliography{neurips_2023}%,iclr2023/anthology,iclr2023/acl2020,iclr2023/iclr2023_conference}

\begin{thebibliography}{74}
\providecommand{\natexlab}[1]{#1}
\providecommand{\url}[1]{\texttt{#1}}
\expandafter\ifx\csname urlstyle\endcsname\relax
  \providecommand{\doi}[1]{doi: #1}\else
  \providecommand{\doi}{doi: \begingroup \urlstyle{rm}\Url}\fi

\bibitem[Abzianidze et~al.(2017)Abzianidze, Bjerva, Evang, Haagsma, van Noord, Ludmann, Nguyen, and Bos]{abzianidze-EtAl:2017:EACLshort}
Abzianidze, L., Bjerva, J., Evang, K., Haagsma, H., van Noord, R., Ludmann, P., Nguyen, D.-D., and Bos, J.
\newblock The parallel meaning bank: Towards a multilingual corpus of translations annotated with compositional meaning representations.
\newblock In \emph{Proceedings of the 15th Conference of the European Chapter of the Association for Computational Linguistics}, EACL~'17, pp.\  242--247, Valencia, Spain, 2017.

\bibitem[Antverg \& Belinkov(2021)Antverg and Belinkov]{antverg2021pitfalls}
Antverg, O. and Belinkov, Y.
\newblock On the pitfalls of analyzing individual neurons in language models.
\newblock In \emph{International Conference on Learning Representations}, 2021.

\bibitem[Antverg \& Belinkov(2022)Antverg and Belinkov]{antverg2022on}
Antverg, O. and Belinkov, Y.
\newblock On the pitfalls of analyzing individual neurons in language models.
\newblock In \emph{International Conference on Learning Representations}, 2022.
\newblock URL \url{https://openreview.net/forum?id=8uz0EWPQIMu}.

\bibitem[Antverg et~al.(2022)Antverg, Ben-David, and Belinkov]{IDANI}
Antverg, O., Ben-David, E., and Belinkov, Y.
\newblock Idani: Inference-time domain adaptation via neuron-level interventions, 2022.
\newblock URL \url{https://arxiv.org/abs/2206.00259}.

\bibitem[Arps et~al.(2022)Arps, Samih, Kallmeyer, and Sajjad]{arps-etal-2022-probing}
Arps, D., Samih, Y., Kallmeyer, L., and Sajjad, H.
\newblock Probing for constituency structure in neural language models.
\newblock In \emph{Findings of the Association for Computational Linguistics: EMNLP 2022}, pp.\  6738--6757, Abu Dhabi, United Arab Emirates, December 2022. Association for Computational Linguistics.
\newblock \doi{10.18653/v1/2022.findings-emnlp.502}.
\newblock URL \url{https://aclanthology.org/2022.findings-emnlp.502}.

\bibitem[Bau et~al.(2019)Bau, Belinkov, Sajjad, Durrani, Dalvi, and Glass]{bau2018identifying}
Bau, A., Belinkov, Y., Sajjad, H., Durrani, N., Dalvi, F., and Glass, J.
\newblock Identifying and controlling important neurons in neural machine translation.
\newblock In \emph{International Conference on Learning Representations}, 2019.
\newblock URL \url{https://openreview.net/forum?id=H1z-PsR5KX}.

\bibitem[Belinkov \& Glass(2019)Belinkov and Glass]{belinkov-glass-2019-analysis}
Belinkov, Y. and Glass, J.
\newblock Analysis methods in neural language processing: A survey.
\newblock \emph{Transactions of the Association for Computational Linguistics}, 7:\penalty0 49--72, March 2019.
\newblock \doi{10.1162/tacl_a_00254}.
\newblock URL \url{https://www.aclweb.org/anthology/Q19-1004}.

\bibitem[Belinkov et~al.(2017)Belinkov, Durrani, Dalvi, Sajjad, and Glass]{belinkov:2017:acl}
Belinkov, Y., Durrani, N., Dalvi, F., Sajjad, H., and Glass, J.
\newblock {What do Neural Machine Translation Models Learn about Morphology?}
\newblock In \emph{Proceedings of the 55th Annual Meeting of the Association for Computational Linguistics (ACL)}, Vancouver, July 2017. Association for Computational Linguistics.
\newblock URL \url{https://aclanthology.coli.uni-saarland.de/pdf/P/P17/P17-1080.pdf}.

\bibitem[Bills et~al.(2023)Bills, Cammarata, Mossing, Tillman, Gao, Goh, Sutskever, Leike, Wu, and Saunders]{bills2023language}
Bills, S., Cammarata, N., Mossing, D., Tillman, H., Gao, L., Goh, G., Sutskever, I., Leike, J., Wu, J., and Saunders, W.
\newblock Language models can explain neurons in language models.
\newblock \url{https://openaipublic.blob.core.windows.net/neuron-explainer/paper/index.html}, 2023.

\bibitem[Bricken et~al.(2023)Bricken, Templeton, Batson, Chen, Jermyn, Conerly, Turner, Anil, Denison, Askell, Lasenby, Wu, Kravec, Schiefer, Maxwell, Joseph, Hatfield-Dodds, Tamkin, Nguyen, McLean, Burke, Hume, Carter, Henighan, and Olah]{bricken2023monosemanticity}
Bricken, T., Templeton, A., Batson, J., Chen, B., Jermyn, A., Conerly, T., Turner, N., Anil, C., Denison, C., Askell, A., Lasenby, R., Wu, Y., Kravec, S., Schiefer, N., Maxwell, T., Joseph, N., Hatfield-Dodds, Z., Tamkin, A., Nguyen, K., McLean, B., Burke, J.~E., Hume, T., Carter, S., Henighan, T., and Olah, C.
\newblock Towards monosemanticity: Decomposing language models with dictionary learning.
\newblock \emph{Transformer Circuits Thread}, 2023.
\newblock https://transformer-circuits.pub/2023/monosemantic-features/index.html.

\bibitem[Colombo et~al.(2022)Colombo, Noiry, Irurozki, and CLEMENCON]{nlp_voting_benchmarking}
Colombo, P., Noiry, N., Irurozki, E., and CLEMENCON, S.
\newblock What are the best systems? new perspectives on {NLP} benchmarking.
\newblock In Oh, A.~H., Agarwal, A., Belgrave, D., and Cho, K. (eds.), \emph{Advances in Neural Information Processing Systems}, 2022.
\newblock URL \url{https://openreview.net/forum?id=kvtVrzQPvgb}.

\bibitem[Conneau et~al.(2018)Conneau, Kruszewski, Lample, Barrault, and Baroni]{conneau2018you}
Conneau, A., Kruszewski, G., Lample, G., Barrault, L., and Baroni, M.
\newblock {What you can cram into a single vector: Probing sentence embeddings for linguistic properties}.
\newblock In \emph{Proceedings of the 56th Annual Meeting of the Association for Computational Linguistics (ACL)}, July 2018.

\bibitem[Conneau et~al.(2019)Conneau, Khandelwal, Goyal, Chaudhary, Wenzek, Guzm{\'{a}}n, Grave, Ott, Zettlemoyer, and Stoyanov]{xlmroberta}
Conneau, A., Khandelwal, K., Goyal, N., Chaudhary, V., Wenzek, G., Guzm{\'{a}}n, F., Grave, E., Ott, M., Zettlemoyer, L., and Stoyanov, V.
\newblock Unsupervised cross-lingual representation learning at scale.
\newblock \emph{CoRR}, abs/1911.02116, 2019.
\newblock URL \url{http://arxiv.org/abs/1911.02116}.

\bibitem[Cooper \& Zillante(2012)Cooper and Zillante]{cumulative_voting}
Cooper, D. and Zillante, A.
\newblock A comparison of cumulative voting and generalized plurality voting.
\newblock In \emph{Public Choice}, 2012.
\newblock \doi{10.1007/s11127-010-9707-5}.

\bibitem[Dai et~al.(2022)Dai, Dong, Hao, Sui, Chang, and Wei]{dai-etal-2022-knowledge}
Dai, D., Dong, L., Hao, Y., Sui, Z., Chang, B., and Wei, F.
\newblock Knowledge neurons in pretrained transformers.
\newblock In \emph{Proceedings of the 60th Annual Meeting of the Association for Computational Linguistics (Volume 1: Long Papers)}, pp.\  8493--8502, Dublin, Ireland, May 2022. Association for Computational Linguistics.
\newblock \doi{10.18653/v1/2022.acl-long.581}.
\newblock URL \url{https://aclanthology.org/2022.acl-long.581}.

\bibitem[Dalvi et~al.(2019)Dalvi, Durrani, Sajjad, Belinkov, Bau, and Glass]{dalvi:2019:AAAI}
Dalvi, F., Durrani, N., Sajjad, H., Belinkov, Y., Bau, D.~A., and Glass, J.
\newblock What is one grain of sand in the desert? analyzing individual neurons in deep nlp models.
\newblock In \emph{Proceedings of the Thirty-Third AAAI Conference on Artificial Intelligence (AAAI, Oral presentation)}, January 2019.

\bibitem[Dalvi et~al.(2020)Dalvi, Sajjad, Durrani, and Belinkov]{dalvi-2020-CCFS}
Dalvi, F., Sajjad, H., Durrani, N., and Belinkov, Y.
\newblock Analyzing redundancy in pretrained transformer models.
\newblock In \emph{Proceedings of the 2020 Conference on Empirical Methods in Natural Language Processing (EMNLP-2020)}, Online, November 2020.

\bibitem[Dalvi et~al.(2022)Dalvi, Khan, Alam, Durrani, Xu, and Sajjad]{dalvi2022discovering}
Dalvi, F., Khan, A.~R., Alam, F., Durrani, N., Xu, J., and Sajjad, H.
\newblock Discovering latent concepts learned in {BERT}.
\newblock In \emph{International Conference on Learning Representations}, 2022.
\newblock URL \url{https://openreview.net/forum?id=POTMtpYI1xH}.

\bibitem[Dalvi et~al.(2023)Dalvi, Durrani, and Sajjad]{dalvi-etal-2023-neurox}
Dalvi, F., Durrani, N., and Sajjad, H.
\newblock Neurox library for neuron analysis of deep nlp models.
\newblock In \emph{Proceedings of the 61st Annual Meeting of the Association for Computational Linguistics: System Demonstrations}, pp.\  75--83, Toronto, Canada, July 2023. Association for Computational Linguistics.

\bibitem[Daniel(1990)]{daniel1990applied}
Daniel, W.
\newblock \emph{Applied Nonparametric Statistics}.
\newblock Duxbury advanced series in statistics and decision sciences. PWS-KENT Pub., 1990.
\newblock ISBN 9780534919764.
\newblock URL \url{https://books.google.ca/books?id=0hPvAAAAMAAJ}.

\bibitem[Danilevsky et~al.(2020)Danilevsky, Qian, Aharonov, Katsis, Kawas, and Sen]{danilevsky-etal-2020-survey}
Danilevsky, M., Qian, K., Aharonov, R., Katsis, Y., Kawas, B., and Sen, P.
\newblock A survey of the state of explainable {AI} for natural language processing.
\newblock In \emph{Proceedings of the 1st Conference of the Asia-Pacific Chapter of the Association for Computational Linguistics and the 10th International Joint Conference on Natural Language Processing}, pp.\  447--459, Suzhou, China, December 2020. Association for Computational Linguistics.
\newblock URL \url{https://www.aclweb.org/anthology/2020.aacl-main.46}.

\bibitem[Denil et~al.(2014)Denil, Demiraj, and de~Freitas]{DBLP:journals/corr/DenilDF14}
Denil, M., Demiraj, A., and de~Freitas, N.
\newblock Extraction of salient sentences from labelled documents.
\newblock \emph{CoRR}, abs/1412.6815, 2014.
\newblock URL \url{http://arxiv.org/abs/1412.6815}.

\bibitem[Devlin et~al.(2019)Devlin, Chang, Lee, and Toutanova]{devlin-etal-2019-bert}
Devlin, J., Chang, M.-W., Lee, K., and Toutanova, K.
\newblock {BERT}: Pre-training of deep bidirectional transformers for language understanding.
\newblock In \emph{Proceedings of the 2019 Conference of the North {A}merican Chapter of the Association for Computational Linguistics: Human Language Technologies, Volume 1 (Long and Short Papers)}, Minneapolis, Minnesota, 2019. Association for Computational Linguistics.

\bibitem[Dhamdhere et~al.(2018)Dhamdhere, Sundararajan, and Yan]{kamdhere}
Dhamdhere, K., Sundararajan, M., and Yan, Q.
\newblock How important is a neuron?
\newblock \emph{CoRR}, abs/1805.12233, 2018.
\newblock URL \url{http://arxiv.org/abs/1805.12233}.

\bibitem[Durrani et~al.(2019)Durrani, Dalvi, Sajjad, Belinkov, and Nakov]{durrani-etal-2019-one}
Durrani, N., Dalvi, F., Sajjad, H., Belinkov, Y., and Nakov, P.
\newblock One size does not fit all: Comparing {NMT} representations of different granularities.
\newblock In \emph{Proceedings of the 2019 Conference of the North {A}merican Chapter of the Association for Computational Linguistics: Human Language Technologies, Volume 1 (Long and Short Papers)}, pp.\  1504--1516, Minneapolis, Minnesota, June 2019. Association for Computational Linguistics.
\newblock \doi{10.18653/v1/N19-1154}.
\newblock URL \url{https://www.aclweb.org/anthology/N19-1154}.

\bibitem[Durrani et~al.(2020)Durrani, Sajjad, Dalvi, and Belinkov]{durrani-etal-2020-analyzing}
Durrani, N., Sajjad, H., Dalvi, F., and Belinkov, Y.
\newblock Analyzing individual neurons in pre-trained language models.
\newblock In \emph{Proceedings of the 2020 Conference on Empirical Methods in Natural Language Processing (EMNLP)}, pp.\  4865--4880, Online, November 2020. Association for Computational Linguistics.
\newblock \doi{10.18653/v1/2020.emnlp-main.395}.
\newblock URL \url{https://www.aclweb.org/anthology/2020.emnlp-main.395}.

\bibitem[Durrani et~al.(2022)Durrani, Dalvi, and Sajjad]{neuronJournal}
Durrani, N., Dalvi, F., and Sajjad, H.
\newblock Linguistic correlation analysis: Discovering salient neurons in deepnlp models, 2022.
\newblock URL \url{https://arxiv.org/abs/2206.13288}.

\bibitem[Elhage et~al.(2021)Elhage, Nanda, Olsson, Henighan, Joseph, Mann, Askell, Bai, Chen, Conerly, DasSarma, Drain, Ganguli, Hatfield-Dodds, Hernandez, Jones, Kernion, Lovitt, Ndousse, Amodei, Brown, Clark, Kaplan, McCandlish, and Olah]{elhage2021mathematical}
Elhage, N., Nanda, N., Olsson, C., Henighan, T., Joseph, N., Mann, B., Askell, A., Bai, Y., Chen, A., Conerly, T., DasSarma, N., Drain, D., Ganguli, D., Hatfield-Dodds, Z., Hernandez, D., Jones, A., Kernion, J., Lovitt, L., Ndousse, K., Amodei, D., Brown, T., Clark, J., Kaplan, J., McCandlish, S., and Olah, C.
\newblock A mathematical framework for transformer circuits.
\newblock \emph{Transformer Circuits Thread}, 2021.
\newblock https://transformer-circuits.pub/2021/framework/index.html.

\bibitem[Elhage et~al.(2022)Elhage, Hume, Olsson, Nanda, Henighan, Johnston, ElShowk, Joseph, DasSarma, Mann, Hernandez, Askell, Ndousse, Drain, Chen, Bai, Ganguli, Lovitt, Hatfield-Dodds, Kernion, Conerly, Kravec, Fort, Kadavath, Jacobson, Tran-Johnson, Kaplan, Clark, Brown, McCandlish, Amodei, and Olah]{elhage2022solu}
Elhage, N., Hume, T., Olsson, C., Nanda, N., Henighan, T., Johnston, S., ElShowk, S., Joseph, N., DasSarma, N., Mann, B., Hernandez, D., Askell, A., Ndousse, K., Drain, D., Chen, A., Bai, Y., Ganguli, D., Lovitt, L., Hatfield-Dodds, Z., Kernion, J., Conerly, T., Kravec, S., Fort, S., Kadavath, S., Jacobson, J., Tran-Johnson, E., Kaplan, J., Clark, J., Brown, T., McCandlish, S., Amodei, D., and Olah, C.
\newblock Softmax linear units.
\newblock \emph{Transformer Circuits Thread}, 2022.
\newblock https://transformer-circuits.pub/2022/solu/index.html.

\bibitem[Ethayarajh(2019)]{ethayarajh-2019-contextual}
Ethayarajh, K.
\newblock How contextual are contextualized word representations? comparing the geometry of {BERT}, {ELM}o, and {GPT}-2 embeddings.
\newblock In \emph{Proceedings of the 2019 Conference on Empirical Methods in Natural Language Processing and the 9th International Joint Conference on Natural Language Processing (EMNLP-IJCNLP)}, pp.\  55--65, Hong Kong, China, November 2019. Association for Computational Linguistics.
\newblock \doi{10.18653/v1/D19-1006}.
\newblock URL \url{https://www.aclweb.org/anthology/D19-1006}.

\bibitem[Foote et~al.(2023)Foote, Nanda, Kran, Konstas, Cohen, and Barez]{foote2023neuron}
Foote, A., Nanda, N., Kran, E., Konstas, I., Cohen, S., and Barez, F.
\newblock Neuron to graph: Interpreting language model neurons at scale, 2023.

\bibitem[Geva et~al.(2021)Geva, Schuster, Berant, and Levy]{geva-etal-2021-transformer}
Geva, M., Schuster, R., Berant, J., and Levy, O.
\newblock Transformer feed-forward layers are key-value memories.
\newblock In \emph{Proceedings of the 2021 Conference on Empirical Methods in Natural Language Processing}, pp.\  5484--5495, Online and Punta Cana, Dominican Republic, November 2021. Association for Computational Linguistics.
\newblock \doi{10.18653/v1/2021.emnlp-main.446}.
\newblock URL \url{https://aclanthology.org/2021.emnlp-main.446}.

\bibitem[Hennigen et~al.(2020)Hennigen, Williams, and Cotterell]{torroba-hennigen-etal-2020-intrinsic}
Hennigen, L.~T., Williams, A., and Cotterell, R.
\newblock Intrinsic probing through dimension selection.
\newblock In \emph{Proceedings of the 2020 Conference on Empirical Methods in Natural Language Processing (EMNLP)}, pp.\  197--216, Online, November 2020. Association for Computational Linguistics.
\newblock \doi{10.18653/v1/2020.emnlp-main.15}.
\newblock URL \url{https://www.aclweb.org/anthology/2020.emnlp-main.15}.

\bibitem[Hewitt \& Liang(2019)Hewitt and Liang]{hewitt-liang-2019-designing}
Hewitt, J. and Liang, P.
\newblock Designing and interpreting probes with control tasks.
\newblock In \emph{Proceedings of the 2019 Conference on Empirical Methods in Natural Language Processing and the 9th International Joint Conference on Natural Language Processing (EMNLP-IJCNLP)}, pp.\  2733--2743, Hong Kong, China, November 2019. Association for Computational Linguistics.
\newblock \doi{10.18653/v1/D19-1275}.
\newblock URL \url{https://aclanthology.org/D19-1275}.

\bibitem[Hupkes et~al.(2018)Hupkes, Veldhoen, and Zuidema]{hupkes2018visualisation}
Hupkes, D., Veldhoen, S., and Zuidema, W.
\newblock Visualisation and 'diagnostic classifiers' reveal how recurrent and recursive neural networks process hierarchical structure, 2018.

\bibitem[K{\'a}d{\'a}r et~al.(2017{\natexlab{a}})K{\'a}d{\'a}r, Chrupa{\l}a, and Alishahi]{kadar-etal-2017-representation}
K{\'a}d{\'a}r, {\'A}., Chrupa{\l}a, G., and Alishahi, A.
\newblock Representation of linguistic form and function in recurrent neural networks.
\newblock \emph{Computational Linguistics}, 43\penalty0 (4):\penalty0 761--780, December 2017{\natexlab{a}}.
\newblock \doi{10.1162/COLI_a_00300}.
\newblock URL \url{https://www.aclweb.org/anthology/J17-4003}.

\bibitem[K{\'a}d{\'a}r et~al.(2017{\natexlab{b}})K{\'a}d{\'a}r, Chrupa{\l}a, and Alishahi]{kadar2016representation}
K{\'a}d{\'a}r, A., Chrupa{\l}a, G., and Alishahi, A.
\newblock Representation of linguistic form and function in recurrent neural networks.
\newblock \emph{Computational Linguistics}, 43\penalty0 (4):\penalty0 761--780, 2017{\natexlab{b}}.

\bibitem[Karpathy et~al.(2015)Karpathy, Johnson, and Fei-Fei]{karpathy2015visualizing}
Karpathy, A., Johnson, J., and Fei-Fei, L.
\newblock Visualizing and understanding recurrent networks.
\newblock \emph{arXiv preprint arXiv:1506.02078}, 2015.

\bibitem[Kendall(1938)]{kendall1938measure}
Kendall, M.
\newblock A new measure of rank correlation.
\newblock \emph{Biometrika}, 1938.

\bibitem[Lakretz et~al.(2019)Lakretz, Kruszewski, Desbordes, Hupkes, Dehaene, and Baroni]{lakretz-etal-2019-emergence}
Lakretz, Y., Kruszewski, G., Desbordes, T., Hupkes, D., Dehaene, S., and Baroni, M.
\newblock The emergence of number and syntax units in {LSTM} language models.
\newblock In \emph{Proceedings of the 2019 Conference of the North {A}merican Chapter of the Association for Computational Linguistics: Human Language Technologies, Volume 1 (Long and Short Papers)}, pp.\  11--20, Minneapolis, Minnesota, June 2019. Association for Computational Linguistics.
\newblock \doi{10.18653/v1/N19-1002}.
\newblock URL \url{https://www.aclweb.org/anthology/N19-1002}.

\bibitem[Lapata(2006)]{lapata-2006-automatic}
Lapata, M.
\newblock Automatic evaluation of information ordering: Kendall{'}s tau.
\newblock \emph{Computational Linguistics}, 32\penalty0 (4):\penalty0 471--484, 2006.
\newblock \doi{10.1162/coli.2006.32.4.471}.
\newblock URL \url{https://aclanthology.org/J06-4002}.

\bibitem[Li et~al.(2016{\natexlab{a}})Li, Chen, Hovy, and Jurafsky]{li-etal-2016-visualizing}
Li, J., Chen, X., Hovy, E., and Jurafsky, D.
\newblock Visualizing and understanding neural models in {NLP}.
\newblock In \emph{Proceedings of the 2016 Conference of the North {A}merican Chapter of the Association for Computational Linguistics: Human Language Technologies}, pp.\  681--691, San Diego, California, June 2016{\natexlab{a}}. Association for Computational Linguistics.
\newblock \doi{10.18653/v1/N16-1082}.
\newblock URL \url{https://www.aclweb.org/anthology/N16-1082}.

\bibitem[Li et~al.(2016{\natexlab{b}})Li, Monroe, and Jurafsky]{erasure_li}
Li, J., Monroe, W., and Jurafsky, D.
\newblock Understanding neural networks through representation erasure.
\newblock \emph{CoRR}, abs/1612.08220, 2016{\natexlab{b}}.
\newblock URL \url{http://arxiv.org/abs/1612.08220}.

\bibitem[Lin et~al.(2017)Lin, Li, and Guo]{consensus_ranking}
Lin, Z., Li, Y., and Guo, X.
\newblock Consensus of rankings.
\newblock \emph{CoRR}, abs/1704.08464, 2017.
\newblock URL \url{http://arxiv.org/abs/1704.08464}.

\bibitem[Lippman(2012)]{voting_book}
Lippman, D.
\newblock \emph{Voting Theory, Math in Society}.
\newblock Pierce College Ft Steilacoom, 2012.

\bibitem[Liu et~al.(2019{\natexlab{a}})Liu, Gardner, Belinkov, Peters, and Smith]{liu-etal-2019-linguistic}
Liu, N.~F., Gardner, M., Belinkov, Y., Peters, M.~E., and Smith, N.~A.
\newblock Linguistic knowledge and transferability of contextual representations.
\newblock In \emph{Proceedings of the 2019 Conference of the North {A}merican Chapter of the Association for Computational Linguistics: Human Language Technologies, Volume 1 (Long and Short Papers)}, pp.\  1073--1094, Minneapolis, Minnesota, June 2019{\natexlab{a}}. Association for Computational Linguistics.
\newblock URL \url{https://www.aclweb.org/anthology/N19-1112}.

\bibitem[Liu et~al.(2019{\natexlab{b}})Liu, Ott, Goyal, Du, Joshi, Chen, Levy, Lewis, Zettlemoyer, and Stoyanov]{roberta}
Liu, Y., Ott, M., Goyal, N., Du, J., Joshi, M., Chen, D., Levy, O., Lewis, M., Zettlemoyer, L., and Stoyanov, V.
\newblock Roberta: {A} robustly optimized {BERT} pretraining approach.
\newblock \emph{CoRR}, abs/1907.11692, 2019{\natexlab{b}}.
\newblock URL \url{http://arxiv.org/abs/1907.11692}.

\bibitem[Lundberg \& Lee(2017)Lundberg and Lee]{shappely_NIPS2017_7062}
Lundberg, S.~M. and Lee, S.-I.
\newblock A unified approach to interpreting model predictions.
\newblock In Guyon, I., Luxburg, U.~V., Bengio, S., Wallach, H., Fergus, R., Vishwanathan, S., and Garnett, R. (eds.), \emph{Advances in Neural Information Processing Systems 30}, pp.\  4765--4774. Curran Associates, Inc., 2017.
\newblock URL \url{http://papers.nips.cc/paper/7062-a-unified-approach-to-interpreting-model-predictions.pdf}.

\bibitem[Marcus et~al.(1993)Marcus, Santorini, and Marcinkiewicz]{marcus-etal-1993-building}
Marcus, M.~P., Santorini, B., and Marcinkiewicz, M.~A.
\newblock Building a large annotated corpus of {E}nglish: The {P}enn {T}reebank.
\newblock \emph{Computational Linguistics}, 19\penalty0 (2):\penalty0 313--330, 1993.
\newblock URL \url{https://www.aclweb.org/anthology/J93-2004}.

\bibitem[Martins \& Astudillo(2016)Martins and Astudillo]{pmlr-v48-martins16}
Martins, A. and Astudillo, R.
\newblock From softmax to sparsemax: A sparse model of attention and multi-label classification.
\newblock In Balcan, M.~F. and Weinberger, K.~Q. (eds.), \emph{Proceedings of The 33rd International Conference on Machine Learning}, volume~48 of \emph{Proceedings of Machine Learning Research}, pp.\  1614--1623, New York, New York, USA, 20--22 Jun 2016. PMLR.
\newblock URL \url{https://proceedings.mlr.press/v48/martins16.html}.

\bibitem[Mousi et~al.(2023)Mousi, Durrani, and Dalvi]{mousi2023llms}
Mousi, B., Durrani, N., and Dalvi, F.
\newblock Can llms facilitate interpretation of pre-trained language models?
\newblock In \emph{Proceedings of the 2023 Conference on Empirical Methods in Natural Language Processing (EMNLP)}, Singapore, dec 2023. Association for Computational Linguistics.

\bibitem[Mu \& Andreas(2020)Mu and Andreas]{Mu-Nips}
Mu, J. and Andreas, J.
\newblock Compositional explanations of neurons.
\newblock \emph{CoRR}, abs/2006.14032, 2020.
\newblock URL \url{https://arxiv.org/abs/2006.14032}.

\bibitem[Na et~al.(2019)Na, Choe, Lee, and Kim]{Na-ICLR}
Na, S., Choe, Y.~J., Lee, D., and Kim, G.
\newblock Discovery of natural language concepts in individual units of {CNNs}.
\newblock \emph{CoRR}, abs/1902.07249, 2019.
\newblock URL \url{http://arxiv.org/abs/1902.07249}.

\bibitem[Nanda et~al.(2023)Nanda, Chan, Lieberum, Smith, and Steinhardt]{nanda2023progress}
Nanda, N., Chan, L., Lieberum, T., Smith, J., and Steinhardt, J.
\newblock Progress measures for grokking via mechanistic interpretability.
\newblock In \emph{The Eleventh International Conference on Learning Representations}, 2023.
\newblock URL \url{https://openreview.net/forum?id=9XFSbDPmdW}.

\bibitem[O'Connor \& Robertson(2003)O'Connor and Robertson]{voting_theory}
O'Connor, J. and Robertson, E.
\newblock The mactutor history of mathematics archive, 2003.

\bibitem[Olsson et~al.(2022)Olsson, Elhage, Nanda, Joseph, DasSarma, Henighan, Mann, Askell, Bai, Chen, Conerly, Drain, Ganguli, Hatfield-Dodds, Hernandez, Johnston, Jones, Kernion, Lovitt, Ndousse, Amodei, Brown, Clark, Kaplan, McCandlish, and Olah]{olsson2022context}
Olsson, C., Elhage, N., Nanda, N., Joseph, N., DasSarma, N., Henighan, T., Mann, B., Askell, A., Bai, Y., Chen, A., Conerly, T., Drain, D., Ganguli, D., Hatfield-Dodds, Z., Hernandez, D., Johnston, S., Jones, A., Kernion, J., Lovitt, L., Ndousse, K., Amodei, D., Brown, T., Clark, J., Kaplan, J., McCandlish, S., and Olah, C.
\newblock In-context learning and induction heads.
\newblock \emph{Transformer Circuits Thread}, 2022.
\newblock https://transformer-circuits.pub/2022/in-context-learning-and-induction-heads/index.html.

\bibitem[Patrão~Neves(2016)]{consensus_inbook}
Patrão~Neves, M.
\newblock \emph{Consensus}, pp.\  19--29.
\newblock Encyclopedia of Global Bioethics, Springer, 06 2016.
\newblock \doi{10.1007/978-3-319-05544-2_119-1}.

\bibitem[Poerner et~al.(2018)Poerner, Roth, and Sch{\"u}tze]{poerner-etal-2018-interpretable}
Poerner, N., Roth, B., and Sch{\"u}tze, H.
\newblock Interpretable textual neuron representations for {NLP}.
\newblock In \emph{Proceedings of the 2018 {EMNLP} Workshop {B}lackbox{NLP}: Analyzing and Interpreting Neural Networks for {NLP}}, pp.\  325--327, Brussels, Belgium, November 2018. Association for Computational Linguistics.
\newblock \doi{10.18653/v1/W18-5437}.
\newblock URL \url{https://www.aclweb.org/anthology/W18-5437}.

\bibitem[Prasanna et~al.(2020)Prasanna, Rogers, and Rumshisky]{prasanna-etal-2020-bert}
Prasanna, S., Rogers, A., and Rumshisky, A.
\newblock {W}hen {BERT} {P}lays the {L}ottery, {A}ll {T}ickets {A}re {W}inning.
\newblock In \emph{Proceedings of the 2020 Conference on Empirical Methods in Natural Language Processing (EMNLP)}, pp.\  3208--3229, Online, November 2020. Association for Computational Linguistics.
\newblock \doi{10.18653/v1/2020.emnlp-main.259}.
\newblock URL \url{https://aclanthology.org/2020.emnlp-main.259}.

\bibitem[Radford et~al.(2019)Radford, Wu, Child, Luan, Amodei, Sutskever, et~al.]{Radford}
Radford, A., Wu, J., Child, R., Luan, D., Amodei, D., Sutskever, I., et~al.
\newblock Language models are unsupervised multitask learners.
\newblock \emph{OpenAI blog}, 1\penalty0 (8):\penalty0 9, 2019.

\bibitem[Ribeiro et~al.(2018)Ribeiro, Singh, and Guestrin]{10.5555/3504035.3504222}
Ribeiro, M.~T., Singh, S., and Guestrin, C.
\newblock Anchors: High-precision model-agnostic explanations.
\newblock In \emph{Proceedings of the Thirty-Second AAAI Conference on Artificial Intelligence and Thirtieth Innovative Applications of Artificial Intelligence Conference and Eighth AAAI Symposium on Educational Advances in Artificial Intelligence}, AAAI'18/IAAI'18/EAAI'18. AAAI Press, 2018.
\newblock ISBN 978-1-57735-800-8.

\bibitem[Sajjad et~al.(2022{\natexlab{a}})Sajjad, Alam, Dalvi, and Durrani]{sajjad-etal-2022-coling-postprocessing}
Sajjad, H., Alam, F., Dalvi, F., and Durrani, N.
\newblock Effect of post-processing on contextualized word representations.
\newblock In \emph{Proceedings of the 29th International Conference on Computational Linguistics}, pp.\  3127--3142, Gyeongju, Republic of Korea, October 2022{\natexlab{a}}. International Committee on Computational Linguistics.
\newblock URL \url{https://aclanthology.org/2022.coling-1.277}.

\bibitem[Sajjad et~al.(2022{\natexlab{b}})Sajjad, Durrani, and Dalvi]{neuronSurvey}
Sajjad, H., Durrani, N., and Dalvi, F.
\newblock {Neuron-level Interpretation of Deep NLP Models: A Survey}.
\newblock \emph{Transactions of the Association for Computational Linguistics}, 2022{\natexlab{b}}.

\bibitem[Suau et~al.(2020)Suau, Zappella, and Apostoloff]{suau2020finding}
Suau, X., Zappella, L., and Apostoloff, N.
\newblock Finding experts in transformer models.
\newblock \emph{CoRR}, abs/2005.07647, 2020.
\newblock URL \url{https://arxiv.org/abs/2005.07647}.

\bibitem[Sundararajan et~al.(2017)Sundararajan, Taly, and Yan]{integrated_gradient}
Sundararajan, M., Taly, A., and Yan, Q.
\newblock Axiomatic attribution for deep networks.
\newblock In \emph{Proceedings of the 34th International Conference on Machine Learning - Volume 70}, ICML'17, pp.\  3319–3328. JMLR.org, 2017.

\bibitem[Tenney et~al.(2019)Tenney, Das, and Pavlick]{tenney-etal-2019-bert}
Tenney, I., Das, D., and Pavlick, E.
\newblock {BERT} rediscovers the classical {NLP} pipeline.
\newblock In \emph{Proceedings of the 57th Annual Meeting of the Association for Computational Linguistics}, pp.\  4593--4601, Florence, Italy, July 2019. Association for Computational Linguistics.
\newblock \doi{10.18653/v1/P19-1452}.
\newblock URL \url{https://www.aclweb.org/anthology/P19-1452}.

\bibitem[Tjong Kim~Sang \& Buchholz(2000)Tjong Kim~Sang and Buchholz]{tjong-kim-sang-buchholz-2000-introduction}
Tjong Kim~Sang, E.~F. and Buchholz, S.
\newblock Introduction to the {C}o{NLL}-2000 shared task chunking.
\newblock In \emph{Fourth Conference on Computational Natural Language Learning and the Second Learning Language in Logic Workshop}, 2000.
\newblock URL \url{https://www.aclweb.org/anthology/W00-0726}.

\bibitem[Vig(2019)]{DBLP:journals/corr/abs-1904-02679}
Vig, J.
\newblock Visualizing attention in transformer-based language representation models.
\newblock \emph{CoRR}, abs/1904.02679, 2019.
\newblock URL \url{http://arxiv.org/abs/1904.02679}.

\bibitem[Voita et~al.(2019)Voita, Talbot, Moiseev, Sennrich, and Titov]{voita-etal-2019-analyzing}
Voita, E., Talbot, D., Moiseev, F., Sennrich, R., and Titov, I.
\newblock Analyzing multi-head self-attention: Specialized heads do the heavy lifting, the rest can be pruned.
\newblock In \emph{Proceedings of the 57th Annual Meeting of the Association for Computational Linguistics}, pp.\  5797--5808, Florence, Italy, July 2019. Association for Computational Linguistics.
\newblock \doi{10.18653/v1/P19-1580}.
\newblock URL \url{https://www.aclweb.org/anthology/P19-1580}.

\bibitem[Wang et~al.(2022{\natexlab{a}})Wang, Variengien, Conmy, Shlegeris, and Steinhardt]{wang2022interpretability}
Wang, K., Variengien, A., Conmy, A., Shlegeris, B., and Steinhardt, J.
\newblock Interpretability in the wild: a circuit for indirect object identification in gpt-2 small, 2022{\natexlab{a}}.

\bibitem[Wang et~al.(2022{\natexlab{b}})Wang, Wen, Zhang, Hou, Liu, and Li]{wang-etal-2022-finding-skill}
Wang, X., Wen, K., Zhang, Z., Hou, L., Liu, Z., and Li, J.
\newblock Finding skill neurons in pre-trained transformer-based language models.
\newblock In \emph{Proceedings of the 2022 Conference on Empirical Methods in Natural Language Processing}, pp.\  11132--11152, Abu Dhabi, United Arab Emirates, December 2022{\natexlab{b}}. Association for Computational Linguistics.
\newblock \doi{10.18653/v1/2022.emnlp-main.765}.
\newblock URL \url{https://aclanthology.org/2022.emnlp-main.765}.

\bibitem[Yilmaz et~al.(2008)Yilmaz, Aslam, and Robertson]{rank_correlation_ir}
Yilmaz, E., Aslam, J.~A., and Robertson, S.
\newblock A new rank correlation coefficient for information retrieval.
\newblock In \emph{Proceedings of the 31st Annual International ACM SIGIR Conference on Research and Development in Information Retrieval}, SIGIR '08, pp.\  587–594, New York, NY, USA, 2008. Association for Computing Machinery.
\newblock ISBN 9781605581644.
\newblock \doi{10.1145/1390334.1390435}.
\newblock URL \url{https://doi.org/10.1145/1390334.1390435}.

\bibitem[Zhang \& Bowman(2018)Zhang and Bowman]{zhang-bowman-2018-language}
Zhang, K. and Bowman, S.
\newblock Language modeling teaches you more than translation does: Lessons learned through auxiliary syntactic task analysis.
\newblock In \emph{Proceedings of the 2018 {EMNLP} Workshop {B}lackbox{NLP}: Analyzing and Interpreting Neural Networks for {NLP}}, pp.\  359--361, Brussels, Belgium, November 2018. Association for Computational Linguistics.
\newblock \doi{10.18653/v1/W18-5448}.
\newblock URL \url{https://www.aclweb.org/anthology/W18-5448}.

\bibitem[Zou \& Hastie(2005)Zou and Hastie]{Zou05regularizationand}
Zou, H. and Hastie, T.
\newblock Regularization and variable selection via the elastic net.
\newblock \emph{Journal of the Royal Statistical Society, Series B}, 67:\penalty0 301--320, 2005.

\end{thebibliography}
\bibliographystyle{icml2023}
%\bibliographystyle{iclr2023/NIPS2023/neurips_2023}

% \newpage
% \input{iclr2023/NIPS2023/rebuttal_neurips}

% \newpage
 \useunder{\uline}{\ul}{}
\newpage
\section{Appendix}

\begin{figure*}[htb]
\centering
\subfigure[BERT]{
\begin{minipage}[t]{0.31\linewidth}
\includegraphics[width=1\textwidth]{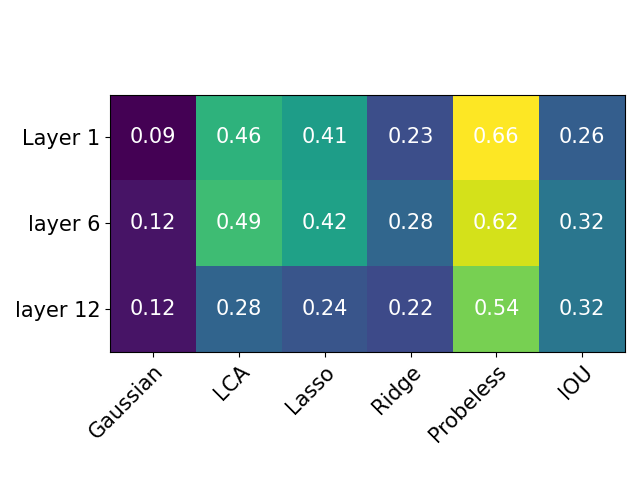}
%\caption{BERT}
\label{fig:selectivityBERT}
\vspace{-2mm}
\end{minipage}
}
\subfigure[RoBERTa]{
\begin{minipage}[t]{0.31\linewidth}
\includegraphics[width=1\textwidth]{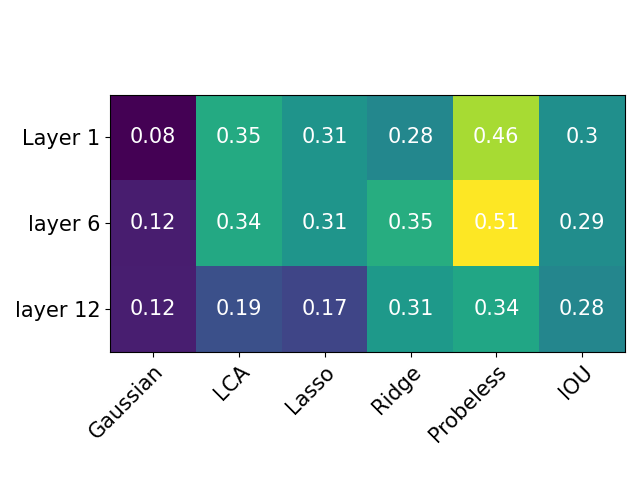}
%\caption{RoBERTa}
\label{fig:selectivityroBERTa}
\vspace{-2mm}
\end{minipage}
}
\subfigure[XLMR]{
\begin{minipage}[t]{0.31\linewidth}
\includegraphics[width=1\textwidth]{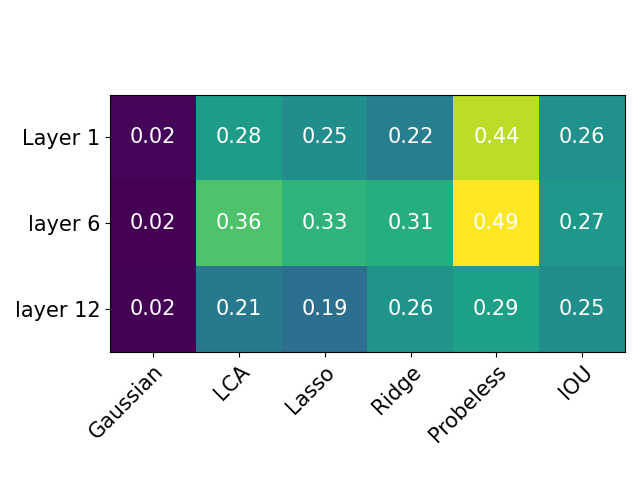}
%\caption{XLMR}
\label{fig:selectivityXLMR}
\vspace{-2mm}
\end{minipage}
}
\centering
\vspace{-2mm}
\caption{\textit{Task: POS,} Pairwise comparison of MeanSelect with other methods using 10--50 neurons}
\label{fig:selectivity_heatmap}
\end{figure*}

\subsection{Case Study}
\label{app:case_study}
\paragraph{Mean Select:} %In this Section, 
We propose a new corpus-based method, \texttt{MeanSelect}, for neuron ranking as a case-study to illustrate %and guide 
how researchers can use our proposed methodology. \cite{kadar2016representation} generated explanations for neurons by extracting top-N 5-gram context for each neuron based on the magnitude of their activations, followed by human annotation. \citet{Na-ICLR} removed the %need for a 
human-in-the-loop by extracting concepts of various granularities from a parsed tree and aligned highly activating neurons to the concepts. Inspired by these works, we propose a novel method, \texttt{MeanSelect}, that generates a ranking of neurons with respect to a concept. The intuition is that a neuron learning a concept will have consistently high activations across different contexts where that concept appears. However, there may be a neuron that always activates with high value irrespective of the concept. A difference in the mean value of the neuron activating the concept and all other concepts will provide the true behavior of the neuron for the concept.
%To filter them out, we The $\mu_-$ term normalizes for neurons that have high activation values irrespective of the particular concept $c$.
Following our %definitions 
notation in Section \ref{sec:methods}, the score of a given neuron $n$ is defined as follows:

\vspace{-2mm}
\begin{equation}
	R(n, \mathcal{C}) = \frac{\mu (\mathcal{C}) - \mu(\hat{\mathcal{C}})}{n_{max} - n_{min}}
	\label{app:eq:selectivity}
\end{equation}

where $\mu(\mathcal{C})$ is the average, $n_{max}$ is the max and $n_{min}$ is the min of %all 
activations $z(n, w)$ where $w \in \mathcal{C}$, and $\mu(\hat{\mathcal{C}})$ is the average of activations over the random concept set.

\paragraph{Compatibility Metrics}
Table~\ref{tab:compatibility_mean_select} shows the compatibility score of \texttt{MeanSelect} using the representation of 3 different layers and compared it with the Random selection of neurons. The %overall 
high compatibility scores show that \texttt{MeanSelect} discovers neurons that are endorsed by other neuron interpretation methods. This serves as a measure of confidence in the proposed %interpretation 
method. Moreover, one may observe that the method has a relatively lower score for the last layer compared to other layers, giving insight into potential improvements that can be made to the behavior of this method.

% \begin{figure}[htb]
% \centering
% \includegraphics[width=1\textwidth]{./figures/sel.png}
% \caption{Selectivity overlap with other methods-BERT}
% \label{fig:selectivity_overlap}
% \end{figure}

% \begin{figure}[htb]
% \centering
% \includegraphics[width=1\textwidth]{./figures/sel_roberta.png}
% \caption{Selectivity overlap with other methods-RoBERTa}
% \label{fig:selectivity_overlap}
% \end{figure}
% \begin{figure}[htb]
% \centering
% \includegraphics[width=1\textwidth]{./figures/sel_xlm_roberta.png}
% \caption{Selectivity overlap with other methods-XLM-R}
% \label{fig:selectivity_overlap}
% \end{figure}

\paragraph{Pairwise Comparison}
The pairwise comparison of methods further provides insights into how the new method relates to other methods in terms of the resulting neurons. Figure~\ref{fig:selectivity_heatmap} shows the heatmaps of three pre-trained models. MeanSelect has the highest overlap with the Probeless method and except Gaussian, it shows an overlap of at least 0.23 points with other methods. While the high overlap of MeanSelect with Probeless is not surprising given both are based on similar intuitions, the overlap with LCA and L1 shows that the method is selecting a diverse set of neurons captured by a variety of methods. Similar to the discussion on compatibility score, here we observe a substantial overlap drop in the last layer and this highlights the potential vulnerability of the method to certain representations.

\subsection{Concept Datasets}
\label{app:datastats}
We consider concepts from three linguistic tasks: parts of speech tags~\citep[POS,][]{marcus-etal-1993-building}, semantic tags
%using the Parallel Meaning Bank data
\citep[SEM,][]{abzianidze-EtAl:2017:EACLshort} and syntactic chunking (Chunking) using CoNLL 2000 shared task dataset \citep{tjong-kim-sang-buchholz-2000-introduction}. For the POS dataset, we used 20 concepts which have a total dataset size of 40137. These concepts include VBG (777), VBZ (908), NNPS (204), DT (4015), TO (1177), CD (1935), JJ (2836), PRP (801), MD (463), RB (1348), VBP (534), VB (1244), NNS (3021), VBN (1082), POS (433), IN (5039), NN (6660), CC (1220), NNP (4698), and VBD (1742).

For the SEM dataset, we used three concepts which have a total dataset size of 120941. These concepts include IST (72240), NOW (24137) and EXS (24564). We used 10 concepts from Chunking which have a total dataset size of 220606. These concepts include B-ADJP (2493), B-ADVP (5081), B-NP (67285), B-PP (26005), B-VP (26078), I-ADJP (805), I-ADVP (532), I-NP (77368) I-PP (339) and I-VP (14620). For all the datasets used in the experiments, we use train/valid/test split 70\%, 15\% and 15\%. 

\subsection{Results } 
\label{app:results}
\subsubsection{Semantic Tagging Concepts}
%Add average results of all models and heatmap of pairwise comparison. Also add some discussion on the same lines of how it is done in the paper but for semantic tags.

For the SEM dataset, we sample 20000 sentences for experimental validation with train/valid/test split 70\%, 15\% and 15\%. We select three tags: IST (intersective), NOW (present tense) and EXS (untensed simple event). Table ~\ref{app:tab:sem:model_comparison} presents the average \neuronvote{} scores across three models. We observed identical trends to that of POS i.e. Probeless is the most consistent method, LCA and Lasso are second best methods but they suffer on the last layers.

%We replicate the experiments and obtain similar tables like Table 2 and Table 3 in our paper. 
\begin{table*}[]
\centering
\footnotesize
\caption{\textit{Task: POS, Model: BERT,} Average \neuronvote{} score of MeanSelect using 10--50 neurons}
\vskip 0.15in

\begin{tabular}{l|ccc|ccc|ccc}
\toprule
            & \multicolumn{3}{c|}{BERT}                        & \multicolumn{3}{c|}{RoBERTa}                     & \multicolumn{3}{c}{XLMR}                         \\
Layers      & 1              & 6              & 12             & 1              & 6              & 12             & 1              & 6              & 12             \\ \midrule
Random      & 0.019    & 0.021    & 0.023    & 0.020    & 0.019    & 0.017    & 0.019    & 0.017    & 0.022    \\
MeanSelect & 0.464 & 0.476 & 0.402 & 0.392 & 0.451 & 0.328 & 0.368 & 0.408 & 0.253 \\
\bottomrule
\end{tabular}
\label{tab:compatibility_mean_select}
\end{table*}

\begin{table*}[]
\centering
\footnotesize
\caption{\textit{Task: Semantic tagging,} Average \neuronvote{} compatibility scores across Semantic tagging concepts when selecting the top 10, 30, and 50 neurons from layers 1, 6 and 12. Bold numbers, underline numbers, and dashed numbers show the first, second, and third best scores respectively}
\vskip 0.15in

\begin{tabular}{l|ccc|ccc|ccc}
\toprule

          & \multicolumn{3}{c|}{BERT}                        & \multicolumn{3}{c|}{RoBERTa}                     & \multicolumn{3}{c}{XLMR}                         \\
Layers    & 1              & 6              & 12             & 1              & 6              & 12             & 1              & 6              & 12             \\ \midrule
Random    & 0.018          & 0.013          & 0.017          & 0.011          & 0.026          & 0.017          & 0.026          & 0.014          & 0.026          \\ \hline
Gaussian  & 0.257          & 0.256          & 0.222          & 0.237          & 0.282          & 0.245          & 0.176          & 0.256          & 0.195          \\
LCA       & \textbf{0.474} & \textbf{0.541} & {\ul 0.385}    & {\ul 0.488}    & {\ul 0.493}    & \dashuline{0.347}          & \dashuline{0.309}          & {\ul 0.455}    & \dashuline{0.367}          \\
Lasso     & \dashuline{0.407}          & \dashuline{0.492}          & 0.322          & 0.391          & 0.460          & 0.328          & 0.294          & 0.396          & 0.358          \\
Ridge     & 0.316          & 0.343          & 0.361          & \dashuline{0.468}          & \dashuline{0.492}          & {\ul 0.575}    & {\ul 0.372}    & \dashuline{0.430}          & \textbf{0.473} \\
Probeless & {\ul 0.450}    & {\ul 0.501}    & \textbf{0.476} & \textbf{0.547} & \textbf{0.586} & \textbf{0.640} & \textbf{0.495} & \textbf{0.571} & {\ul 0.464}    \\
IoU       & 0.344          & 0.332          & \dashuline{0.380}          & 0.288          & 0.287          & 0.242          & 0.277          & 0.262          & 0.282          \\ \bottomrule

\end{tabular}

\label{app:tab:sem:model_comparison}
\end{table*}
\subsection{Chunking Concepts}
%Add average results of all models and heatmap of pairwise comparison. Also add some discussion on the same lines of how it is done in the paper but for chunking
\begin{table*}[]
\centering
\footnotesize
\caption{\textit{Task: Chunking,} Average \neuronvote{} compatibility scores across Chunking concepts when selecting the top 10, 30, and 50 neurons from layers 1, 6 and 12. Bold numbers, underline numbers, and dashed numbers show the first, second, and third best scores respectively}
\vskip 0.15in

\begin{tabular}{c|ccc|ccc|ccc}
\toprule
          & \multicolumn{3}{c|}{BERT}                        & \multicolumn{3}{c|}{RoBERTa}                     & \multicolumn{3}{c}{XLMR}                         \\
Layers    & 1              & 6              & 12             & 1              & 6              & 12             & 1              & 6              & 12             \\ \midrule
Random    & 0.018          & 0.017          & 0.023          & 0.022          & 0.018          & 0.022          & 0.023          & 0.019          & 0.014          \\ \hline
Gaussian  & 0.122          & 0.181          & 0.174          & 0.111          & 0.163          & 0.164          & 0.110          & 0.121          & 0.143          \\
LCA       & \dashuline{0.395}          & \dashuline{0.469}          & 0.300          & {\ul 0.440}    & {\ul 0.422}    & \dashuline{0.328}          & \dashuline{0.336}          & {\ul 0.447}    & 0.388          \\
    Lasso     & {\ul 0.396}    & {\ul 0.472}    & {\ul 0.301}    & \dashuline{0.399}          & \dashuline{0.395}          & 0.322          & {\ul 0.366}    & \dashuline{0.425}          & 0.395          \\
Ridge     & 0.235          & 0.255          & 0.256          & 0.303          & 0.330          & {\ul 0.386}    & 0.254          & 0.289          & {\ul 0.410}    \\
Probeless & \textbf{0.465} & \textbf{0.506} & \textbf{0.422} & \textbf{0.502} & \textbf{0.500} & \textbf{0.514} & \textbf{0.499} & \textbf{0.507} & \textbf{0.463} \\
IoU       & 0.346          & 0.361          & \dashuline{0.321}          & 0.285          & 0.298          & 0.244          & 0.319          & 0.283          & 0.352          \\ \bottomrule
\end{tabular}

\label{app:tab:chunking:model_comparison}
\end{table*}

\begin{figure*}[htb]
\centering
\subfigure[BERT]{
\begin{minipage}[t]{0.32\linewidth}
\includegraphics[width=1\textwidth]{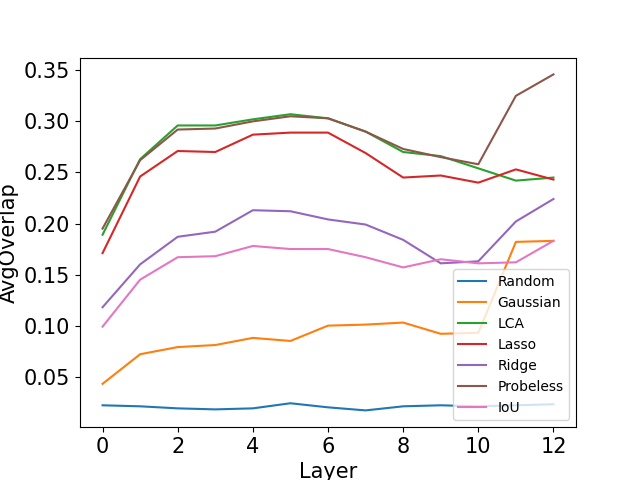}
%\caption{DT Layer 0}
\label{fig:bert_a}
\vspace{-2mm}
\end{minipage}%
}%
\subfigure[RoBERTa]{
\begin{minipage}[t]{0.32\linewidth}
\includegraphics[width=1\textwidth]{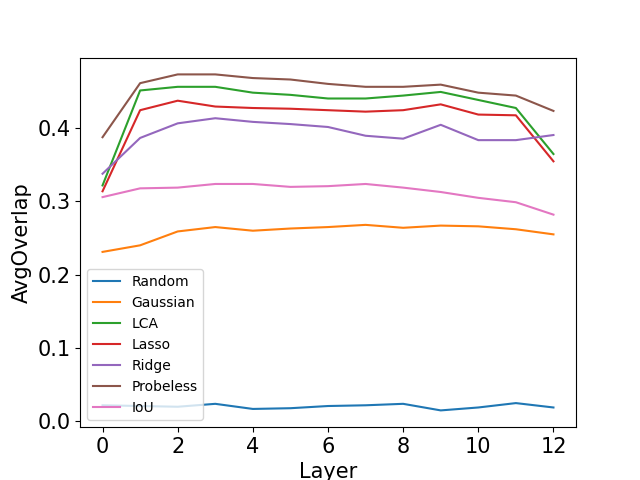}
%\caption{DT Layer 0}
\label{fig:roberta_a}
\vspace{-2mm}
\end{minipage}%
}%
\subfigure[XLMR]{
\begin{minipage}[t]{0.32\linewidth}
\includegraphics[width=1\textwidth]{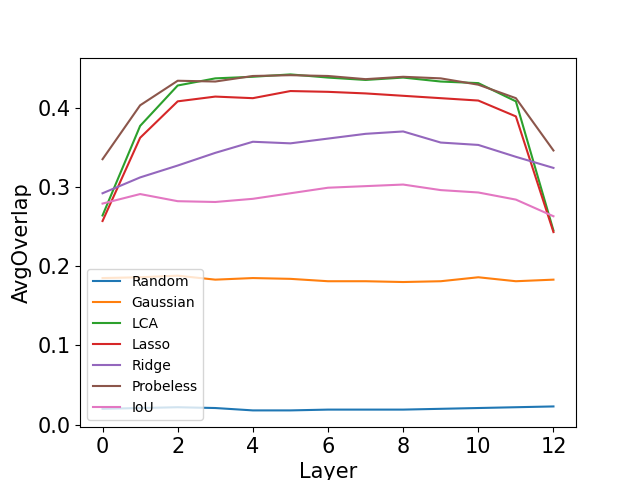}
%\caption{DT Layer 0}
\label{fig:xlmr_a}
\vspace{-2mm}
\end{minipage}%
}%
\vspace{-2mm}

\centering
\caption{\avgoverlap{} score across different Layers in different models. All methods perform better than the baseline, Random. Probeless is the most consistent method across all models, concepts and across all layers. It is among the top methods with LCA and Lasso. However, LCA and Lasso show low score on last layers. The performance of Gaussian deteriorates significantly and is closer to Random when applied on XLMR. }
\label{fig:appendix_avg_overlap}
\end{figure*}
\begin{figure*}[htb]
\centering
\subfigure[BERT]{
\begin{minipage}[t]{0.32\linewidth}
\includegraphics[width=1\textwidth]{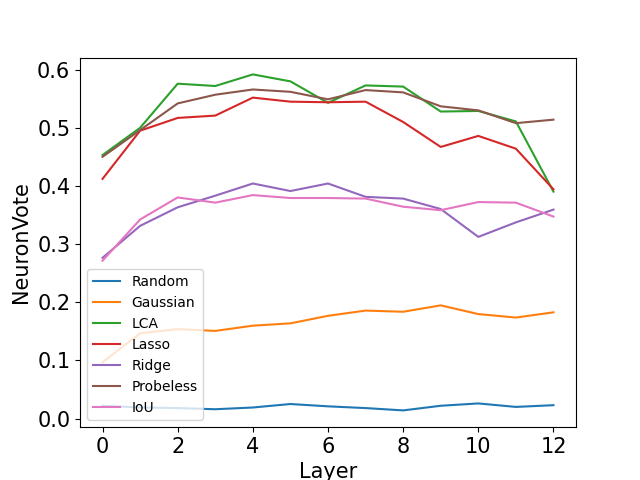}
%\caption{DT Layer 0}
\label{fig:bert_n}
\vspace{-2mm}
\end{minipage}%
}%
\subfigure[RoBERTa]{
\begin{minipage}[t]{0.32\linewidth}
\includegraphics[width=1\textwidth]{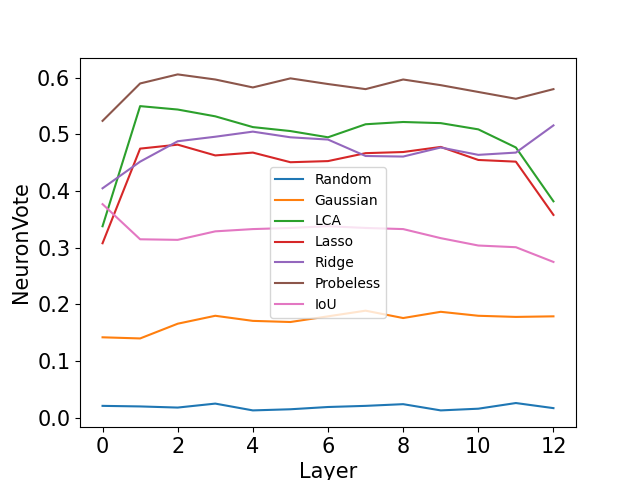}
%\caption{DT Layer 0}
\label{fig:roberta_n}
\vspace{-2mm}
\end{minipage}%
}%
\subfigure[XLMR]{
\begin{minipage}[t]{0.32\linewidth}
\includegraphics[width=1\textwidth]{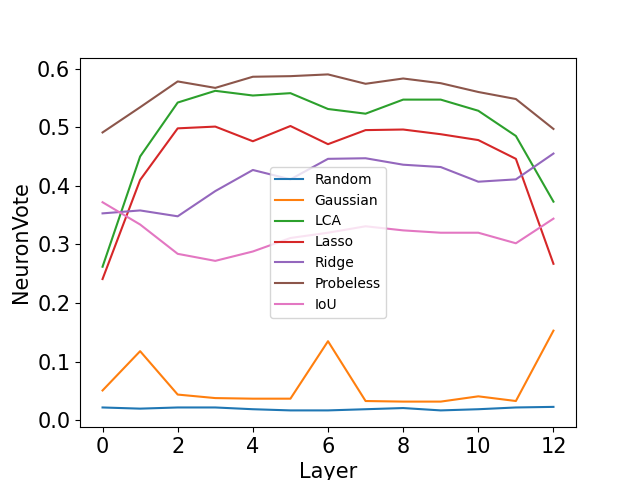}
%\caption{DT Layer 0}
\label{fig:xlmr_n}
\vspace{-2mm}
\end{minipage}%
}%

\centering
\caption{\neuronvote{} score across different Layers in different models. All methods perform better than the baseline, Random. Probeless is the most consistent method across all models, concepts and across all layers. It is among the top methods with LCA and Lasso. However, LCA and Lasso show low score on last layers.}
\label{fig:appendix_neuronvote}
\end{figure*}

Figures \ref{fig:appendix_avg_overlap} and \ref{fig:appendix_neuronvote} show layer-wise results for the two voting methods proposed in the paper, across the three understudied models (BERT, RoBERTa and XLM-R). The results show that voting methods consistently rank the \emph{Probeless} method as the most compatible in terms of neuron rankings across the layers. We are including detailed results in Tables 5--10 give detailed results with exact numbers.

\begin{table*}[]
\centering
\footnotesize
\scriptsize
\caption{This is an extension of Table.\ref{tab:model_comparison}. Average \avgoverlap{} compatibility scores across concepts when selecting the top 10, 30, and 50 neurons. Bold numbers show the best scores. Probeless, LCA are the top performing methods. However, Probeless is most consistent performing method. LCA drops substantially for the last layers.}
\vskip 0.15in

\begin{tabular}{c|ccccccccccccc}
\hline
          & \multicolumn{13}{c}{BERT}                                                                             \\
Layers    & 0     & 1     & 2     & 3     & 4     & 5     & 6     & 7     & 8     & 9     & 10    & 11    & 12    \\ \hline
Random    & 0.022 & 0.021 & 0.019 & 0.018 & 0.019 & 0.024 & 0.020 & 0.017 & 0.021 & 0.022 & 0.021 & 0.022 & 0.023 \\ \hline
Gaussian  & 0.043 & 0.072 & 0.079 & 0.081 & 0.088 & 0.085 & 0.100 & 0.101 & 0.103 & 0.092 & 0.093 & 0.182 & 0.183 \\
LCA       & 0.189 & \textbf{0.263} & \textbf{0.296} & \textbf{0.296} & \textbf{0.302} & \textbf{0.307} & \textbf{0.303} & \textbf{0.290} & 0.27  & \textbf{0.266} & 0.254 & 0.242 & 0.245 \\
Lasso     & 0.171 & 0.246 & 0.271 & 0.270 & 0.287 & 0.289 & 0.289 & 0.269 & 0.245 & 0.247 & 0.24  & 0.253 & 0.243 \\
Ridge     & 0.118 & 0.16  & 0.187 & 0.192 & 0.213 & 0.212 & 0.204 & 0.199 & 0.184 & 0.161 & 0.163 & 0.202 & 0.224 \\
Probeless & \textbf{0.195} & 0.262 & 0.292 & 0.293 & 0.3   & 0.305 & \textbf{0.303} & \textbf{0.29}  & \textbf{0.273} & \textbf{0.265} & \textbf{0.258} & \textbf{0.325} & \textbf{0.346} \\
IoU       & 0.099 & 0.145 & 0.167 & 0.168 & 0.178 & 0.175 & 0.175 & 0.167 & 0.157 & 0.165 & 0.161 & 0.162 & 0.183 \\ \hline
\end{tabular}

\label{app:tab:model_comparison_extend_bert}
\end{table*}
\begin{table*}[]
\centering
\footnotesize
\scriptsize
\caption{This is an extension of Table.\ref{tab:model_comparison}. Average \neuronvote{} compatibility scores across concepts when selecting the top 10, 30, and 50 neurons. 
%Bold numbers, underline numbers, and dashed numbers show the first, second, and third best scores respectively
}
\vskip 0.15in

\begin{tabular}{c|ccccccccccccc}
\hline
          & \multicolumn{13}{c}{BERT}                                                                             \\
Layers    & 0     & 1     & 2     & 3     & 4     & 5     & 6     & 7     & 8     & 9     & 10    & 11    & 12    \\ \hline
Random    & 0.022 & 0.019 & 0.018 & 0.016 & 0.019 & 0.025 & 0.021 & 0.018 & 0.014 & 0.022 & 0.026 & 0.020 & 0.023 \\ \hline
Gaussian  & 0.097 & 0.147 & 0.154 & 0.151 & 0.160 & 0.164 & 0.177 & 0.186 & 0.184 & 0.195 & 0.180 & 0.174 & 0.183 \\
LCA       & \textbf{0.454} & \textbf{0.501} & \textbf{0.577} & \textbf{0.573} & \textbf{0.593} & \textbf{0.581} & 0.544 & \textbf{0.574} & \textbf{0.572} & 0.529 & \textbf{0.530} & \textbf{0.512} & 0.391 \\
Lasso     & 0.413 & 0.496 & 0.518 & 0.522 & 0.553 & 0.546 & 0.545 & 0.546 & 0.511 & 0.468 & 0.487 & 0.465 & 0.395 \\
Ridge     & 0.277 & 0.332 & 0.364 & 0.384 & 0.405 & 0.392 & 0.405 & 0.382 & 0.379 & 0.361 & 0.313 & 0.338 & 0.360 \\
Probeless & 0.451 & 0.497 & 0.543 & 0.558 & 0.567 & 0.563 & \textbf{0.550} & 0.566 & 0.562 & \textbf{0.538} & \textbf{0.531} & 0.509 & 0.515 \\
IoU       & 0.272 & 0.343 & 0.381 & 0.372 & 0.385 & 0.380 & 0.380 & 0.379 & 0.365 & 0.359 & 0.373 & 0.372 & 0.348 \\ \hline
\end{tabular}

\label{app:tab:model_comparison_extend_bert2}
\end{table*}

\begin{table*}[b]
\centering
\footnotesize
\scriptsize
\caption{This is an extension of Table.\ref{tab:model_comparison}. Average \avgoverlap{} compatibility scores across concepts when selecting the top 10, 30, and 50 neurons. 
%Bold numbers, underline numbers, and dashed numbers show the first, second, and third best scores respectively
}
\vskip 0.15in
\begin{tabular}{c|ccccccccccccc}
\hline
          & \multicolumn{13}{c}{RoBERTa}                                                                          \\
Layers    & 0     & 1     & 2     & 3     & 4     & 5     & 6     & 7     & 8     & 9     & 10    & 11    & 12    \\ \hline
Random    & 0.021 & 0.020 & 0.019 & 0.023 & 0.016 & 0.017 & 0.020 & 0.021 & 0.023 & 0.014 & 0.018 & 0.024 & 0.018 \\ \hline
Gaussian  & 0.231 & 0.240 & 0.259 & 0.265 & 0.260 & 0.263 & 0.265 & 0.268 & 0.264 & 0.267 & 0.266 & 0.262 & 0.255 \\
LCA       & 0.322 & 0.452 & 0.457 & 0.457 & 0.449 & 0.446 & 0.441 & 0.441 & 0.445 & 0.450 & 0.439 & 0.428 & 0.365 \\
Lasso     & 0.314 & 0.425 & 0.438 & 0.430 & 0.428 & 0.427 & 0.425 & 0.423 & 0.425 & 0.433 & 0.419 & 0.418 & 0.355 \\
Ridge     & 0.338 & 0.387 & 0.407 & 0.414 & 0.409 & 0.406 & 0.402 & 0.390 & 0.386 & 0.405 & 0.384 & 0.384 & 0.391 \\
Probeless & \textbf{0.388} & \textbf{0.462} & \textbf{0.474} & \textbf{0.474} & \textbf{0.469} & \textbf{0.467} & \textbf{0.461} & \textbf{0.457} & \textbf{0.457} & \textbf{0.460} & \textbf{0.449} & \textbf{0.445} & \textbf{0.424} \\
IoU       & 0.306 & 0.318 & 0.319 & 0.324 & 0.324 & 0.320 & 0.321 & 0.324 & 0.319 & 0.313 & 0.305 & 0.299 & 0.282 \\ \hline
\end{tabular}

\end{table*}
\begin{table*}[b]
\centering
\footnotesize
\scriptsize
\caption{This is an extension of Table.\ref{tab:model_comparison}. Average \neuronvote{} compatibility scores across concepts when selecting the top 10, 30, and 50 neurons. }
\vskip 0.15in

\begin{tabular}{c|ccccccccccccc}
\hline
          & \multicolumn{13}{c}{RoBERTa}                                                                          \\
Layers    & 0     & 1     & 2     & 3     & 4     & 5     & 6     & 7     & 8     & 9     & 10    & 11    & 12    \\ \hline
Random    & 0.021 & 0.020 & 0.018 & 0.025 & 0.013 & 0.015 & 0.019 & 0.021 & 0.024 & 0.013 & 0.016 & 0.026 & 0.017 \\ \hline
Gaussian  & 0.142 & 0.140 & 0.166 & 0.180 & 0.171 & 0.169 & 0.179 & 0.189 & 0.176 & 0.187 & 0.180 & 0.178 & 0.179 \\
LCA       & 0.338 & 0.550 & 0.544 & 0.532 & 0.513 & 0.506 & 0.495 & 0.518 & 0.522 & 0.520 & 0.509 & 0.477 & 0.382 \\
Lasso     & 0.308 & 0.475 & 0.482 & 0.463 & 0.468 & 0.451 & 0.453 & 0.467 & 0.469 & 0.478 & 0.455 & 0.452 & 0.358 \\
Ridge     & 0.405 & 0.452 & 0.488 & 0.496 & 0.505 & 0.495 & 0.491 & 0.462 & 0.461 & 0.477 & 0.464 & 0.468 & 0.516 \\
Probeless & \textbf{0.524} & \textbf{0.590} & \textbf{0.606} & \textbf{0.597} & \textbf{0.583} & \textbf{0.599} & \textbf{0.589} & \textbf{0.580} & \textbf{0.597} & \textbf{0.587} & \textbf{0.575} & \textbf{0.563} & \textbf{0.580} \\
IoU       & 0.377 & 0.315 & 0.314 & 0.329 & 0.333 & 0.335 & 0.338 & 0.335 & 0.333 & 0.317 & 0.304 & 0.301 & 0.275 \\ \hline
\end{tabular}

\label{app:tab:model_comparison_extend_roberta}
\end{table*}

\begin{table*}[b]
\centering
\footnotesize
\scriptsize
\caption{This is an extension of Table.\ref{tab:model_comparison}. Average \avgoverlap{} compatibility scores across concepts when selecting the top 10, 30, and 50 neurons.}
\vskip 0.15in
\begin{tabular}{c|ccccccccccccc}
\hline
          & \multicolumn{13}{c}{XLMR}                                                                             \\
Layers    & 0     & 1     & 2     & 3     & 4     & 5     & 6     & 7     & 8     & 9     & 10    & 11    & 12    \\ \hline
Random    & 0.020 & 0.021 & 0.022 & 0.021 & 0.018 & 0.018 & 0.019 & 0.019 & 0.019 & 0.020 & 0.021 & 0.022 & 0.023 \\ \hline
Gaussian  & 0.185 & 0.186 & 0.188 & 0.183 & 0.185 & 0.184 & 0.181 & 0.181 & 0.180 & 0.181 & 0.186 & 0.181 & 0.183 \\
LCA       & 0.264 & 0.377 & 0.428 & \textbf{0.437} & \textbf{0.439} & \textbf{0.442} & 0.438 & \textbf{0.435} & \textbf{0.438} & 0.433 & \textbf{0.431} & 0.408 & 0.245 \\
Lasso     & 0.257 & 0.362 & 0.408 & 0.414 & 0.412 & 0.421 & 0.420 & 0.418 & 0.415 & 0.412 & 0.409 & 0.389 & 0.243 \\
Ridge     & 0.292 & 0.312 & 0.327 & 0.343 & 0.357 & 0.355 & 0.361 & 0.367 & 0.370 & 0.356 & 0.353 & 0.338 & 0.324 \\
Probeless & \textbf{0.335} & \textbf{0.403} & \textbf{0.434} & 0.433 & \textbf{0.440} & \textbf{0.441} & \textbf{0.440} & \textbf{0.436} & \textbf{0.439} & \textbf{0.437} & 0.429 & \textbf{0.412} & \textbf{0.346} \\
IoU       & 0.279 & 0.291 & 0.282 & 0.281 & 0.285 & 0.292 & 0.299 & 0.301 & 0.303 & 0.296 & 0.293 & 0.284 & 0.263 \\ \hline
\end{tabular}

\label{app:tab:model_comparison_extend_xlmr}
\end{table*}
\begin{table*}[]
\centering
\footnotesize
\scriptsize
\caption{This is an extension of Table.\ref{tab:model_comparison}. Average \neuronvote{} compatibility scores across concepts when selecting the top 10, 30, and 50 neurons.}
\vskip 0.15in

\begin{tabular}{c|ccccccccccccc}
\hline
          & \multicolumn{13}{c}{XLMR}                                                                             \\
Layers    & 0     & 1     & 2     & 3     & 4     & 5     & 6     & 7     & 8     & 9     & 10    & 11    & 12    \\ \hline
Random    & 0.022 & 0.020 & 0.022 & 0.022 & 0.019 & 0.017 & 0.017 & 0.019 & 0.021 & 0.017 & 0.019 & 0.022 & 0.023 \\ \hline
Gaussian  & 0.051 & 0.118 & 0.044 & 0.038 & 0.037 & 0.037 & 0.135 & 0.033 & 0.032 & 0.032 & 0.041 & 0.033 & 0.153 \\
LCA       & 0.262 & 0.450  & 0.542 & 0.562 & 0.554 & 0.558 & 0.531 & 0.523 & 0.547 & 0.547 & 0.528 & 0.485 & 0.373 \\
Lasso     & 0.241 & 0.410  & 0.498 & 0.501 & 0.476 & 0.502 & 0.471 & 0.495 & 0.496 & 0.488 & 0.478 & 0.446 & 0.267 \\
Ridge     & 0.353 & 0.358 & 0.348 & 0.391 & 0.427 & 0.411 & 0.446 & 0.447 & 0.436 & 0.432 & 0.407 & 0.411 & 0.455 \\
Probeless & \textbf{0.491} & \textbf{0.534} & \textbf{0.578} & \textbf{0.567} & \textbf{0.586} & \textbf{0.587} & \textbf{0.590}  & \textbf{0.574} & \textbf{0.583} & \textbf{0.575} & \textbf{0.560} & \textbf{0.548} & \textbf{0.497} \\
IoU       & 0.372 & 0.334 & 0.284 & 0.272 & 0.288 & 0.311 & 0.320  & 0.331 & 0.324 & 0.320 & 0.320 & 0.302 & 0.344 \\ \hline
\end{tabular}
\label{app:tab:model_comparison_extend_xlmr2}
\end{table*}

\begin{table*}[b]
\centering
\footnotesize
\scriptsize
\caption{This is an extension of Table.\ref{tab:compatibility_mean_select}. Average \avgoverlap{} score of MeanSelect when selecting 10, 30, and 50 neurons for all layers}
\vskip 0.15in

\begin{tabular}{c|ccccccccccccc}
\hline
Layers     & 0     & 1     & 2     & 3     & 4     & 5     & 6     & 7     & 8     & 9     & 10    & 11    & 12    \\ \hline
           & \multicolumn{13}{c}{BERT}                                                                             \\
Random     & 0.022 & 0.021 & 0.019 & 0.018 & 0.019 & 0.022 & 0.020 & 0.017 & 0.021 & 0.022 & 0.021 & 0.022 & 0.023 \\
MeanSelect & 0.266 & 0.351 & 0.363 & 0.362 & 0.365 & 0.369 & 0.374 & 0.370 & 0.356 & 0.349 & 0.343 & 0.327 & 0.283 \\ \hline
           & \multicolumn{13}{c}{RoBERTa}                                                                          \\
Random     & 0.021 & 0.020 & 0.019 & 0.023 & 0.016 & 0.017 & 0.020 & 0.021 & 0.023 & 0.014 & 0.018 & 0.024 & 0.018 \\
MeanSelect & 0.252 & 0.294 & 0.307 & 0.317 & 0.326 & 0.328 & 0.323 & 0.320 & 0.314 & 0.308 & 0.283 & 0.269 & 0.239 \\ \hline
           & \multicolumn{13}{c}{XLMR}                                                                             \\
Random     & 0.020 & 0.021 & 0.022 & 0.021 & 0.018 & 0.018 & 0.019 & 0.019 & 0.019 & 0.020 & 0.021 & 0.022 & 0.023 \\
MeanSelect & 0.233 & 0.246 & 0.245 & 0.247 & 0.264 & 0.273 & 0.294 & 0.292 & 0.281 & 0.261 & 0.259 & 0.241 & 0.159 \\ \hline
\end{tabular}
\label{app:tab:compatibility_mean_select_full}
\end{table*}

\begin{table*}[]
\centering
\scriptsize
\caption{This is an extension of Table.\ref{tab:compatibility_mean_select}. Average \neuronvote{} score of MeanSelect when selecting 10, 30, and 50 neurons for all layers}
\vskip 0.15in

\begin{tabular}{c|ccccccccccccc}
\hline
Layers     & 0     & 1     & 2     & 3     & 4     & 5     & 6     & 7     & 8     & 9     & 10    & 11    & 12    \\ \hline
           & \multicolumn{13}{c}{BERT}                                                                             \\
Random     & 0.022 & 0.019 & 0.018 & 0.016 & 0.019 & 0.025 & 0.021 & 0.018 & 0.014 & 0.022 & 0.026 & 0.020 & 0.023 \\
MeanSelect & 0.361 & 0.476 & 0.462 & 0.458 & 0.470 & 0.468 & 0.479 & 0.488 & 0.486 & 0.463 & 0.457 & 0.432 & 0.400 \\ \hline
           & \multicolumn{13}{c}{RoBERTa}                                                                          \\
Random     & 0.021 & 0.020 & 0.018 & 0.025 & 0.013 & 0.015 & 0.019 & 0.021 & 0.024 & 0.013 & 0.016 & 0.026 & 0.017 \\
MeanSelect & 0.358 & 0.388 & 0.395 & 0.403 & 0.426 & 0.423 & 0.455 & 0.438 & 0.437 & 0.406 & 0.383 & 0.374 & 0.326 \\ \hline
           & \multicolumn{13}{c}{XLMR}                                                                             \\
Random     & 0.022 & 0.020  & 0.022 & 0.022 & 0.019 & 0.017 & 0.017 & 0.019 & 0.021 & 0.017 & 0.019 & 0.022 & 0.023 \\
MeanSelect & 0.376 & 0.357 & 0.356 & 0.342 & 0.364 & 0.392 & 0.414 & 0.416 & 0.400 & 0.372 & 0.381 & 0.360 & 0.243 \\ \hline
\end{tabular}
\label{app:tab:compatibility_mean_select_full2}
\end{table*}

\begin{figure*}[htb]
\centering
% \subfigure[CD L0]{
% \begin{minipage}[t]{0.25\linewidth}
% \includegraphics[width=1\textwidth]{./figures/heatmap_layerCD_tag0_neuron_10.png}
% %\caption{DT Layer 0}
% \end{minipage}%
% }%
\subfigure[Layer 1]{
\begin{minipage}[t]{0.32\linewidth}
\includegraphics[width=1\textwidth]{./figures/heatmap_layer1_avg_neuron_bert_1.png}
%\caption{DT Layer 0}
\label{fig:cdl1-bert}
\end{minipage}%
}%
\subfigure[Layer 2]{
\begin{minipage}[t]{0.32\linewidth}
\includegraphics[width=1\textwidth]{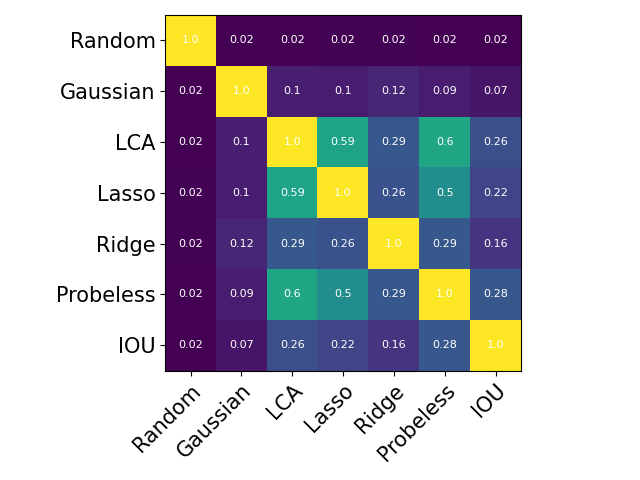}
%\caption{DT Layer 0}
\label{fig:cdl6-bert}
\end{minipage}%
}%
\subfigure[Layer 3]{
\begin{minipage}[t]{0.32\linewidth}
\includegraphics[width=1\textwidth]{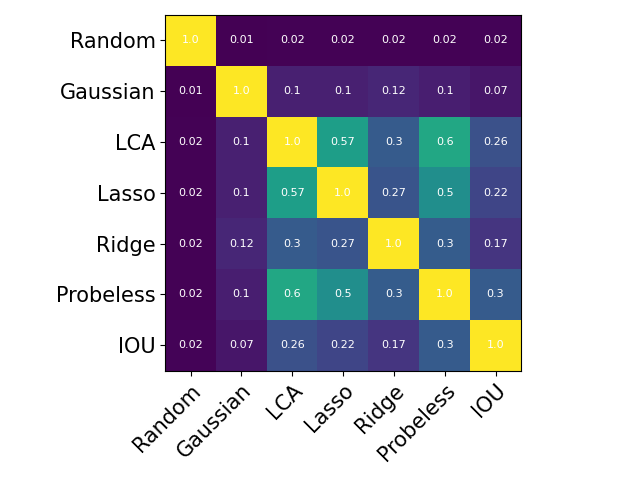}
%\caption{DT Layer 0}
\label{fig:cdl12}
\end{minipage}%
}%
\\
% \subfigure[DT L0]{
% \begin{minipage}[t]{0.25\linewidth}
% \includegraphics[width=1\textwidth]{./figures/heatmap_layerDT_tag0_neuron_10.png}
% \end{minipage}%
% }%
\subfigure[Layer 4]{
\begin{minipage}[t]{0.32\linewidth}
\includegraphics[width=1\textwidth]{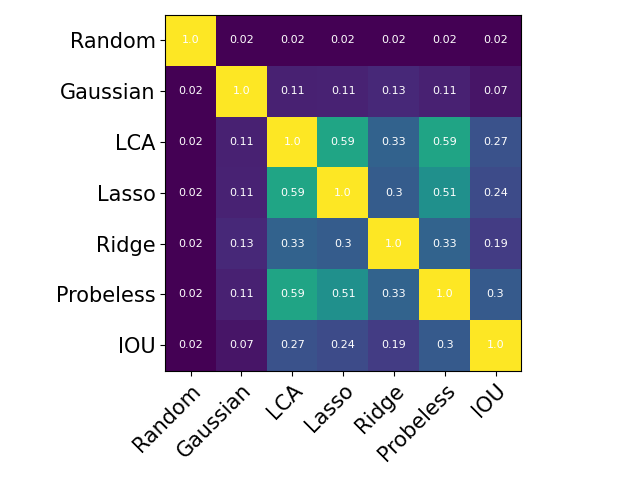}
\label{fig:DTl1-bert}
%\caption{DT Layer 0}
\end{minipage}%
}%
\subfigure[Layer 5]{
\begin{minipage}[t]{0.32\linewidth}
\includegraphics[width=1\textwidth]{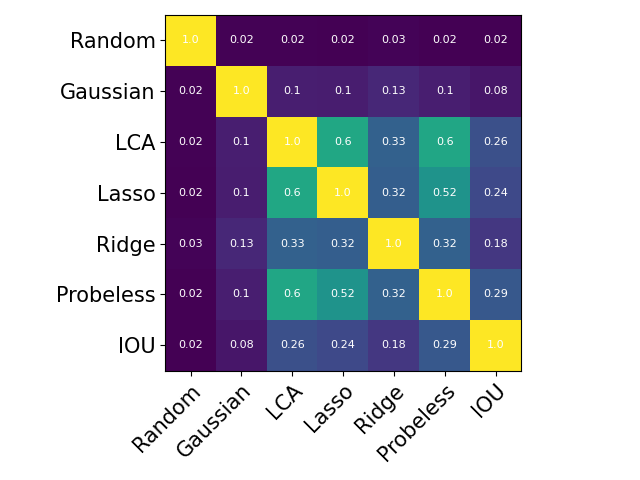}
\label{fig:DTl6-bert}
%\caption{DT Layer 0}
\end{minipage}%
}%
\subfigure[Layer 6]{
\begin{minipage}[t]{0.32\linewidth}
\includegraphics[width=1\textwidth]{./figures/heatmap_layer6_avg_neuron_bert_1.png}
\label{fig:DTl62-bert}
%\caption{DT Layer 0}
\end{minipage}%
}%
\\
% \subfigure[NNPS L0]{
% \begin{minipage}[t]{0.25\linewidth}
% \includegraphics[width=1\textwidth]{./figures/heatmap_layerNNPS_tag0_neuron_10.png}
% \end{minipage}
% }%
\subfigure[Layer 7]{
\begin{minipage}[t]{0.32\linewidth}
\includegraphics[width=1\textwidth]{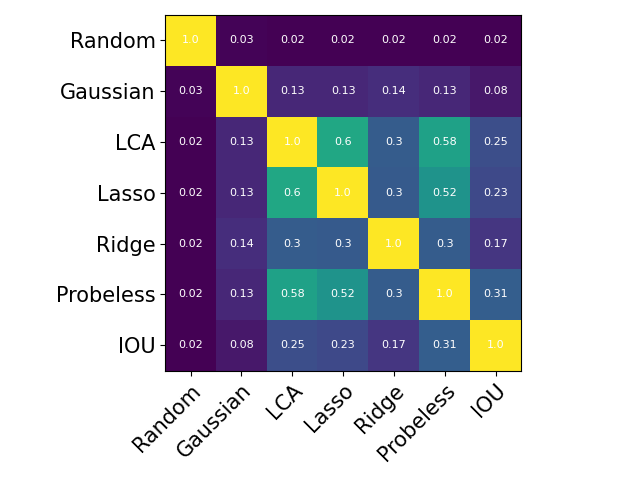}
%\caption{DT Layer 0}
\label{fig:NNPSl61-bert}
\end{minipage}%
}%
\subfigure[Layer 8]{
\begin{minipage}[t]{0.32\linewidth}
\includegraphics[width=1\textwidth]{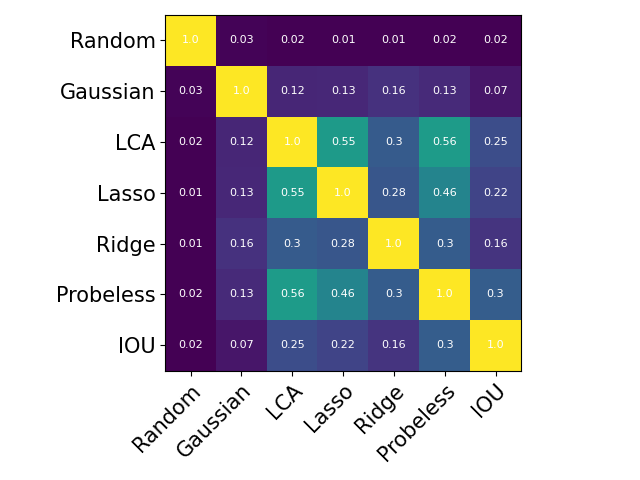}
%\caption{DT Layer 0}
\label{fig:NNPSl66-bert}
\end{minipage}%
}%
\subfigure[Layer 9]{
\begin{minipage}[t]{0.32\linewidth}
\includegraphics[width=1\textwidth]{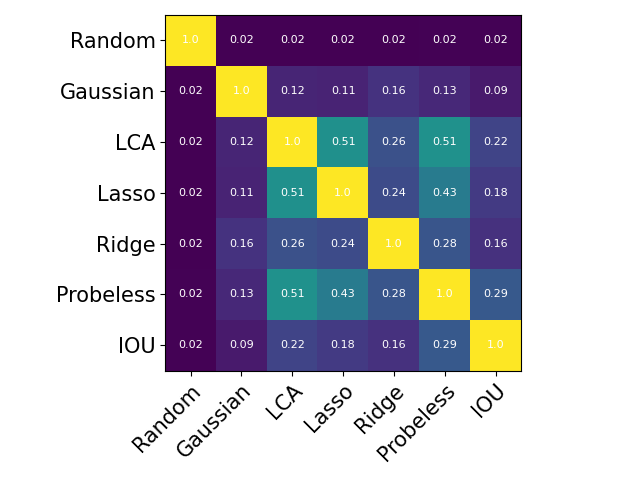}
%\caption{DT Layer 0}
\label{fig:NNPSl612-bert}
\end{minipage}%
}%
\\
% \subfigure[VBZ L0]{
% \begin{minipage}[t]{0.25\linewidth}
% \includegraphics[width=1\textwidth]{./figures/heatmap_layerVBZ_tag0_neuron_10.png}
% \end{minipage}
% }%
\subfigure[Layer 10]{
\begin{minipage}[t]{0.32\linewidth}
\includegraphics[width=1\textwidth]{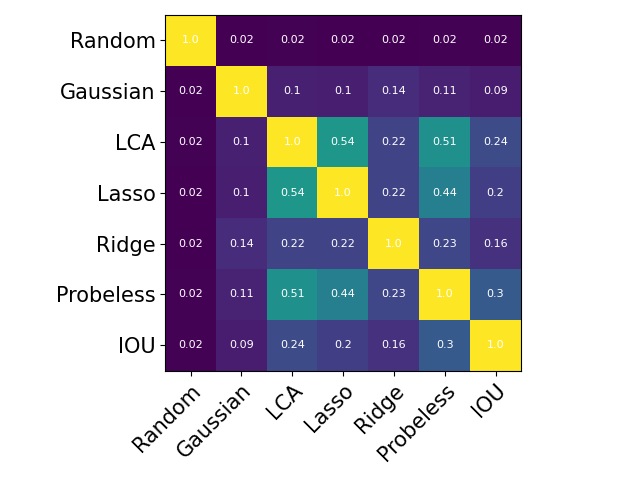}
%\caption{DT Layer 0}
\label{fig:VBZl6-bert}
\end{minipage}%
}%
\subfigure[Layer 11]{
\begin{minipage}[t]{0.32\linewidth}
\includegraphics[width=1\textwidth]{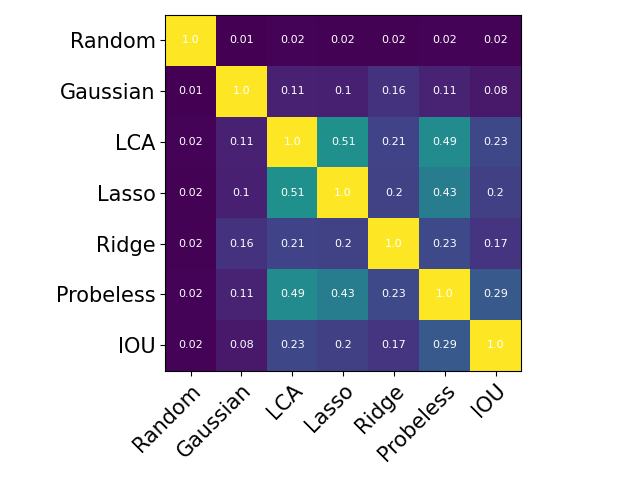}
%\caption{DT Layer 0}
\label{fig:VBZl6}
\end{minipage}%
}%
\subfigure[Layer 12]{
\begin{minipage}[t]{0.32\linewidth}
\includegraphics[width=1\textwidth]{./figures/heatmap_layer12_avg_neuron_bert_1.png}
%\caption{DT Layer 0}
\label{fig:VBZl62-bert}
\end{minipage}%
}%
\\
% \subfigure[VBZ L0]{
% \begin{minipage}[t]{0.25\linewidth}
% \includegraphics[width=1\textwidth]{./figures/heatmap_layerVBZ_tag0_neuron_10.png}
% \end{minipage}
% }%
% \subfigure[NN layer 1]{
% \begin{minipage}[t]{0.32\linewidth}
% \includegraphics[width=1\textwidth]{./figures/heatmap_layer1_NN_neuron_10.png}
% %\caption{DT Layer 0}
% \label{fig:fig:NNPSl661}
% \end{minipage}%
% }%
% \subfigure[NN layer 6]{
% \begin{minipage}[t]{0.32\linewidth}
% \includegraphics[width=1\textwidth]{./figures/heatmap_layer6_NN_neuron_10.png}
% %\caption{DT Layer 0}
% \label{fig:fig:NNPSl666}
% \end{minipage}%
% }%
% \subfigure[NN layer 12]{
% \begin{minipage}[t]{0.32\linewidth}
% \includegraphics[width=1\textwidth]{./figures/heatmap_layer12_NN_neuron_10.png}
% %\caption{DT Layer 0}
% \label{fig:fig:NNPSl6612}
% \end{minipage}%
% }
\centering
\caption{This is an extension of Figure.\ref{fig:top10_50_heatmap}. Comparing average overlap of top 10-50 neurons across methods for BERT}
\label{app:fig:top10_heatmap}
\end{figure*}

\begin{figure*}[htb]
\centering
% \subfigure[CD L0]{
% \begin{minipage}[t]{0.25\linewidth}
% \includegraphics[width=1\textwidth]{./figures/heatmap_layerCD_tag0_neuron_10.png}
% %\caption{DT Layer 0}
% \end{minipage}%
% }%
\subfigure[Layer 1]{
\begin{minipage}[t]{0.32\linewidth}
\includegraphics[width=1\textwidth]{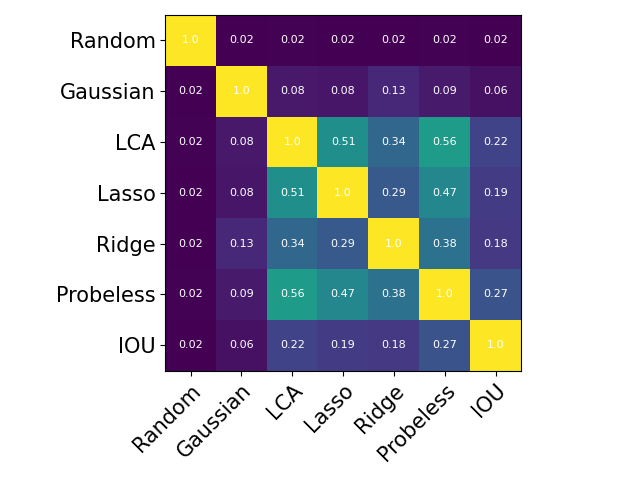}
%\caption{DT Layer 0}
\label{fig:cdl1-roberta}
\end{minipage}%
}%
\subfigure[Layer 2]{
\begin{minipage}[t]{0.32\linewidth}
\includegraphics[width=1\textwidth]{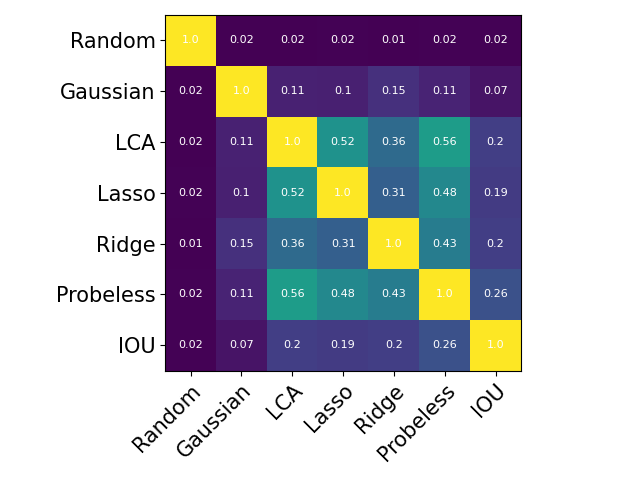}
%\caption{DT Layer 0}
\label{fig:cdl12l6-roberta}
\end{minipage}%
}%
\subfigure[Layer 3]{
\begin{minipage}[t]{0.32\linewidth}
\includegraphics[width=1\textwidth]{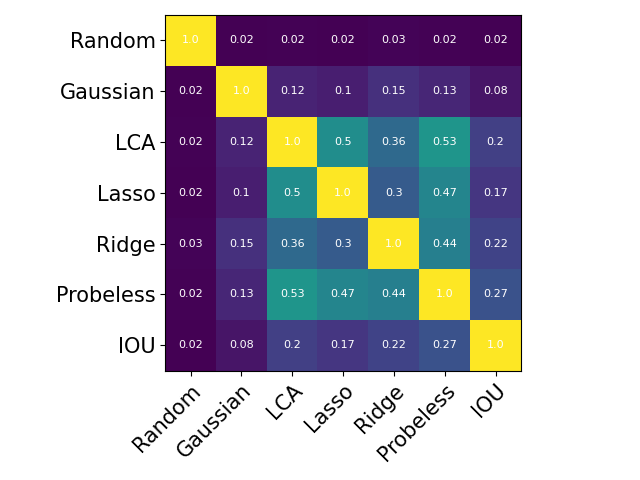}
%\caption{DT Layer 0}
\label{fig:cdl12-roberta}
\end{minipage}%
}%
\\
% \subfigure[DT L0]{
% \begin{minipage}[t]{0.25\linewidth}
% \includegraphics[width=1\textwidth]{./figures/heatmap_layerDT_tag0_neuron_10.png}
% \end{minipage}%
% }%
\subfigure[Layer 4]{
\begin{minipage}[t]{0.32\linewidth}
\includegraphics[width=1\textwidth]{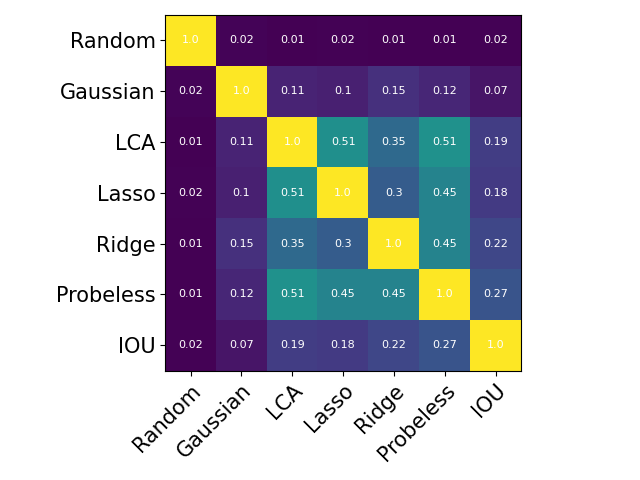}
\label{fig:DTl1-roberta}
%\caption{DT Layer 0}
\end{minipage}%
}%
\subfigure[Layer 5]{
\begin{minipage}[t]{0.32\linewidth}
\includegraphics[width=1\textwidth]{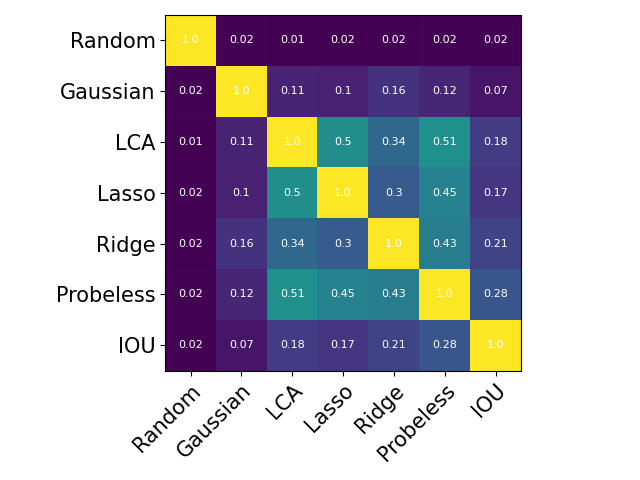}
\label{fig:DTl6-roberta}
%\caption{DT Layer 0}
\end{minipage}%
}%
\subfigure[Layer 6]{
\begin{minipage}[t]{0.32\linewidth}
\includegraphics[width=1\textwidth]{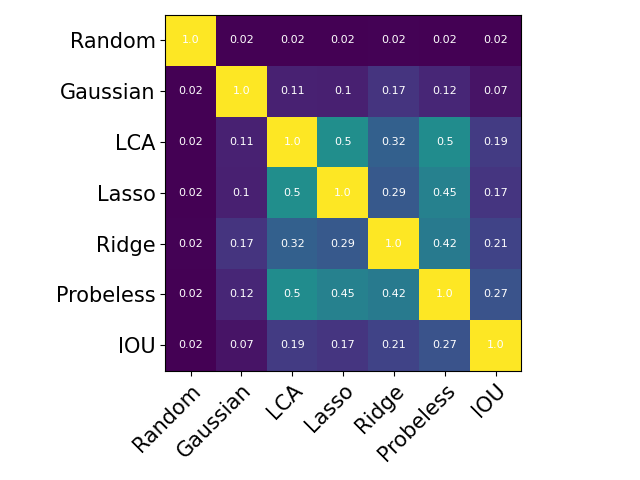}
\label{fig:DTl62-roberta}
%\caption{DT Layer 0}
\end{minipage}%
}%
\\
% \subfigure[NNPS L0]{
% \begin{minipage}[t]{0.25\linewidth}
% \includegraphics[width=1\textwidth]{./figures/heatmap_layerNNPS_tag0_neuron_10.png}
% \end{minipage}
% }%
\subfigure[Layer 7]{
\begin{minipage}[t]{0.32\linewidth}
\includegraphics[width=1\textwidth]{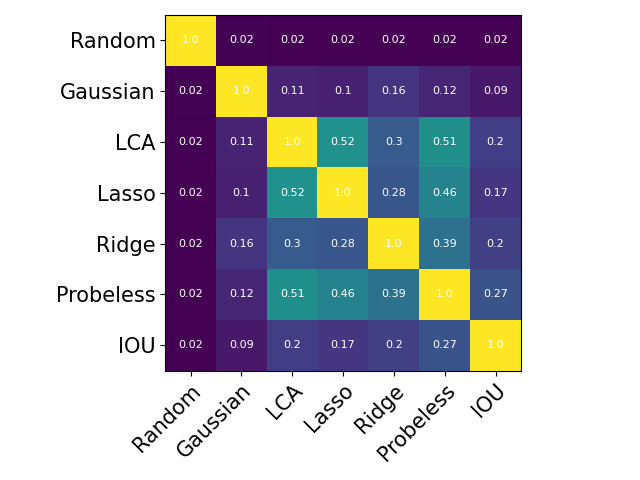}
%\caption{DT Layer 0}
\label{fig:NNPSl61-roberta}
\end{minipage}%
}%
\subfigure[Layer 8]{
\begin{minipage}[t]{0.32\linewidth}
\includegraphics[width=1\textwidth]{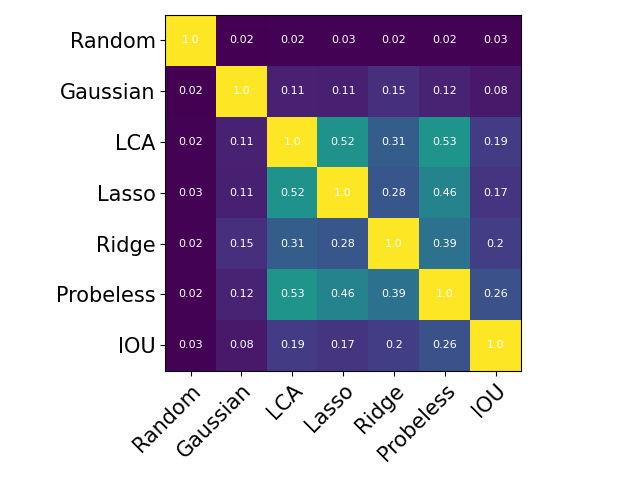}
%\caption{DT Layer 0}
\label{fig:NNPSl66}
\end{minipage}%
}%
\subfigure[Layer 9]{
\begin{minipage}[t]{0.32\linewidth}
\includegraphics[width=1\textwidth]{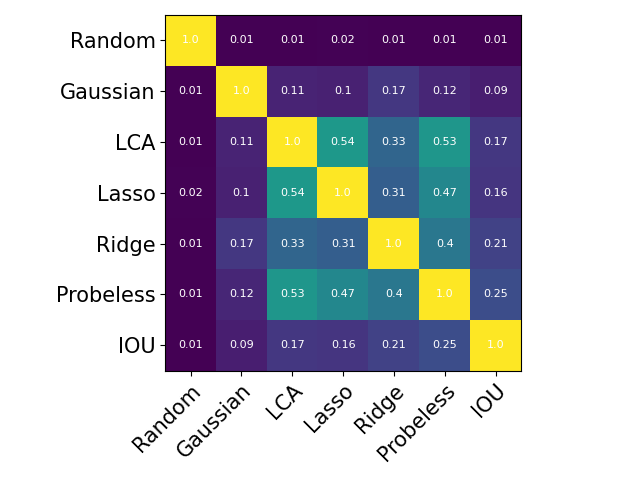}
%\caption{DT Layer 0}
\label{fig:NNPSl12-roberta}
\end{minipage}%
}%
\\
% \subfigure[VBZ L0]{
% \begin{minipage}[t]{0.25\linewidth}
% \includegraphics[width=1\textwidth]{./figures/heatmap_layerVBZ_tag0_neuron_10.png}
% \end{minipage}
% }%
\subfigure[Layer 10]{
\begin{minipage}[t]{0.32\linewidth}
\includegraphics[width=1\textwidth]{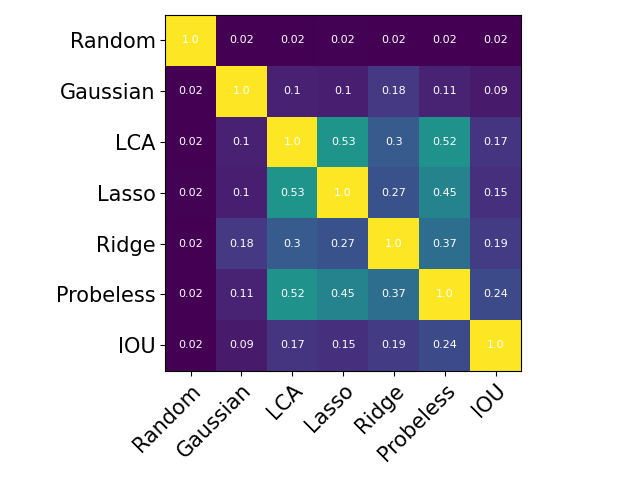}
%\caption{DT Layer 0}
\label{fig:VBZl6-roberta-l10}
\end{minipage}%
}%
\subfigure[Layer 11]{
\begin{minipage}[t]{0.32\linewidth}
\includegraphics[width=1\textwidth]{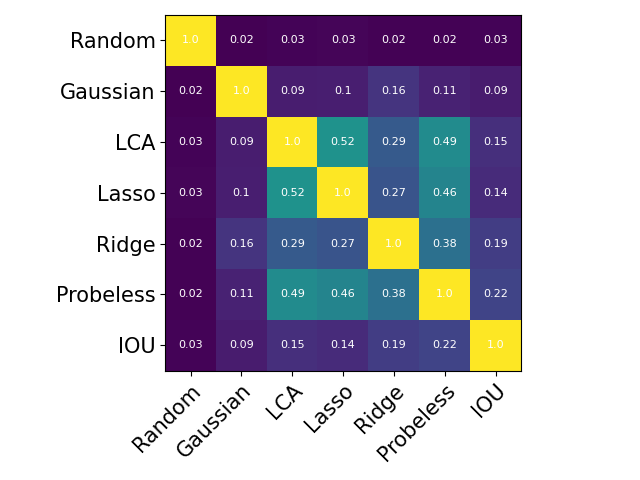}
%\caption{DT Layer 0}
\label{fig:VBZl6-roberta-l11}
\end{minipage}%
}%
\subfigure[Layer 12]{
\begin{minipage}[t]{0.32\linewidth}
\includegraphics[width=1\textwidth]{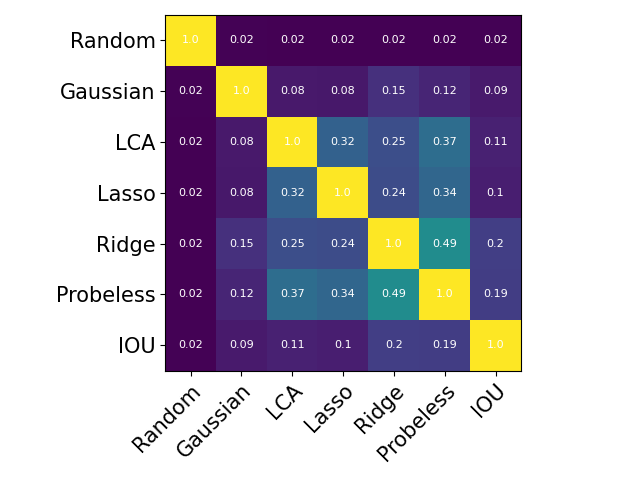}
%\caption{DT Layer 0}
\label{fig:VBZl62-roberta}
\end{minipage}%
}%
\\
% \subfigure[VBZ L0]{
% \begin{minipage}[t]{0.25\linewidth}
% \includegraphics[width=1\textwidth]{./figures/heatmap_layerVBZ_tag0_neuron_10.png}
% \end{minipage}
% }%
% \subfigure[NN layer 1]{
% \begin{minipage}[t]{0.32\linewidth}
% \includegraphics[width=1\textwidth]{./figures/heatmap_layer1_NN_neuron_10.png}
% %\caption{DT Layer 0}
% \label{fig:fig:NNPSl61}
% \end{minipage}%
% }%
% \subfigure[NN layer 6]{
% \begin{minipage}[t]{0.32\linewidth}
% \includegraphics[width=1\textwidth]{./figures/heatmap_layer6_NN_neuron_10.png}
% %\caption{DT Layer 0}
% \label{fig:fig:NNPSl66}
% \end{minipage}%
% }%
% \subfigure[NN layer 12]{
% \begin{minipage}[t]{0.32\linewidth}
% \includegraphics[width=1\textwidth]{./figures/heatmap_layer12_NN_neuron_10.png}
% %\caption{DT Layer 0}
% \label{fig:NNl12}
% \end{minipage}%
% }
\centering
\caption{This is an extension of Figure.\ref{fig:top10_50_heatmap}. Comparing average overlap of top 10-50 neurons across methods for RoBERTa}
\label{app:fig:top10_heatmap_roberta}
\end{figure*}

\begin{figure*}[htb]
\centering
% \subfigure[CD L0]{
% \begin{minipage}[t]{0.25\linewidth}
% \includegraphics[width=1\textwidth]{./figures/heatmap_layerCD_tag0_neuron_10.png}
% %\caption{DT Layer 0}
% \end{minipage}%
% }%
\subfigure[Layer 1]{
\begin{minipage}[t]{0.32\linewidth}
\includegraphics[width=1\textwidth]{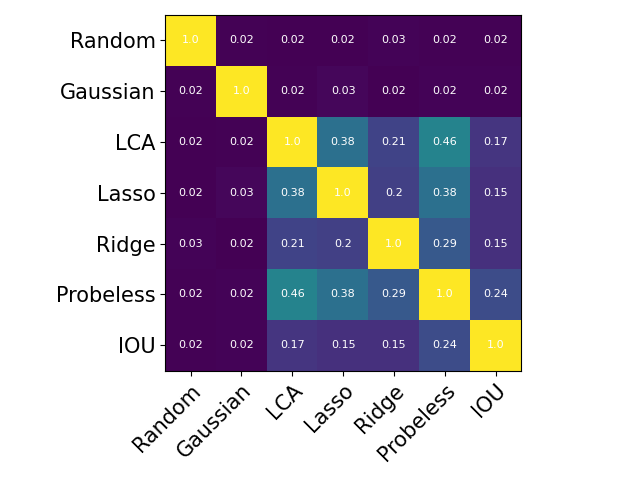}
%\caption{DT Layer 0}
\label{fig:cdl12l1-xlmr}
\end{minipage}%
}%
\subfigure[Layer 2]{
\begin{minipage}[t]{0.32\linewidth}
\includegraphics[width=1\textwidth]{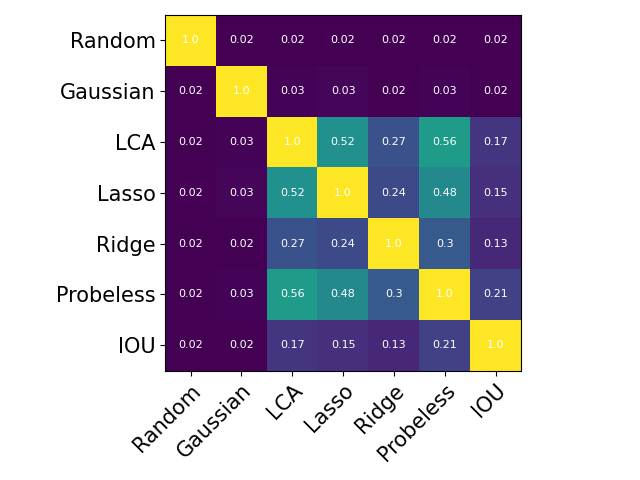}
%\caption{DT Layer 0}
\label{fig:cdl12l6-xlmr}
\end{minipage}%
}%
\subfigure[Layer 3]{
\begin{minipage}[t]{0.32\linewidth}
\includegraphics[width=1\textwidth]{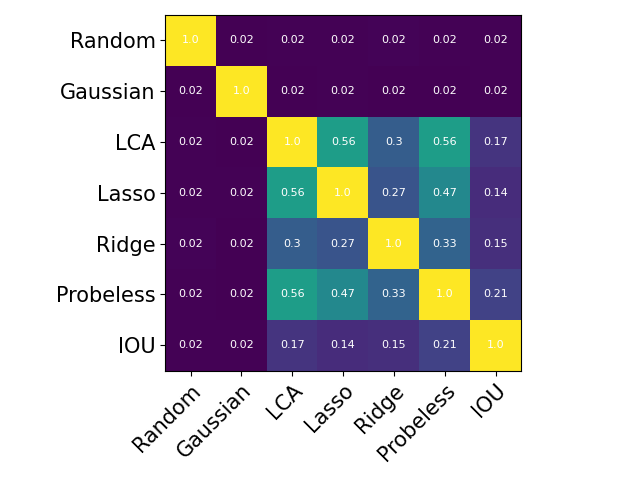}
%\caption{DT Layer 0}
\label{fig:cdl12l12}
\end{minipage}%
}%
\\
% \subfigure[DT L0]{
% \begin{minipage}[t]{0.25\linewidth}
% \includegraphics[width=1\textwidth]{./figures/heatmap_layerDT_tag0_neuron_10.png}
% \end{minipage}%
% }%
\subfigure[Layer 4]{
\begin{minipage}[t]{0.32\linewidth}
\includegraphics[width=1\textwidth]{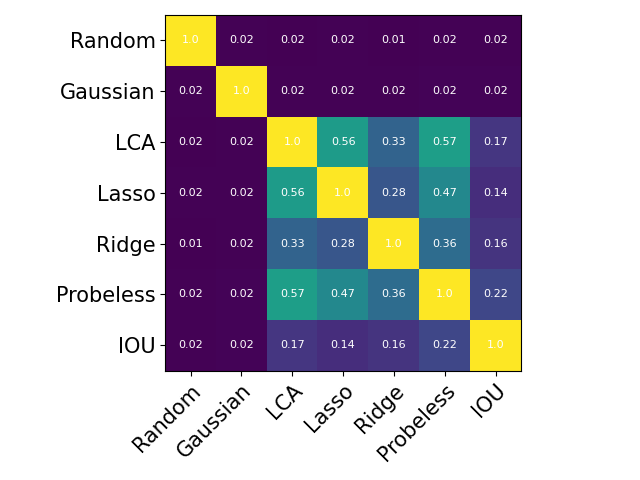}
\label{fig:DTl6-xlmr}
%\caption{DT Layer 0}
\end{minipage}%
}%
\subfigure[Layer 5]{
\begin{minipage}[t]{0.32\linewidth}
\includegraphics[width=1\textwidth]{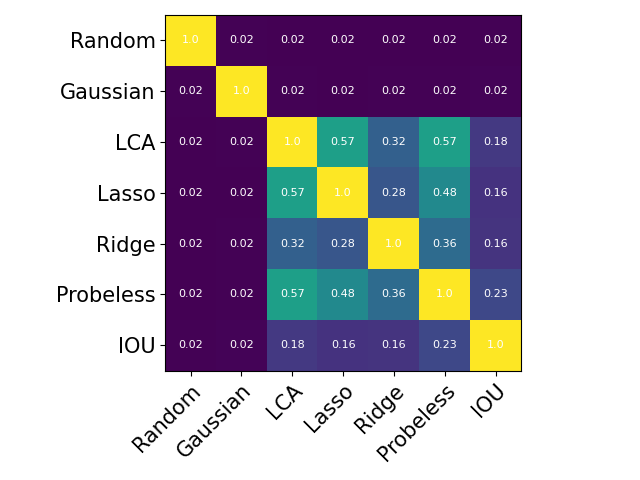}
\label{fig:DTl6}
%\caption{DT Layer 0}
\end{minipage}%
}%
\subfigure[Layer 6]{
\begin{minipage}[t]{0.32\linewidth}
\includegraphics[width=1\textwidth]{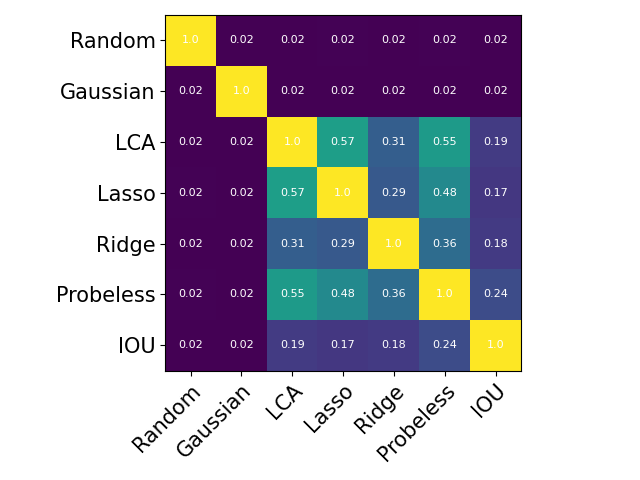}
\label{fig:DTl62-xlmr}
%\caption{DT Layer 0}
\end{minipage}%
}%
\\
% \subfigure[NNPS L0]{
% \begin{minipage}[t]{0.25\linewidth}
% \includegraphics[width=1\textwidth]{./figures/heatmap_layerNNPS_tag0_neuron_10.png}
% \end{minipage}
% }%
\subfigure[Layer 7]{
\begin{minipage}[t]{0.32\linewidth}
\includegraphics[width=1\textwidth]{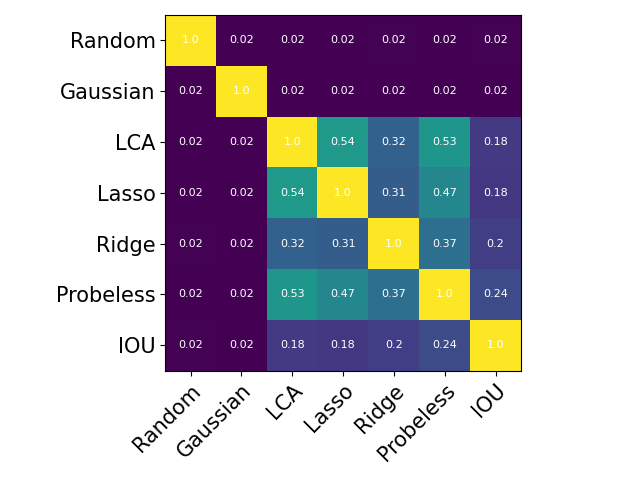}
%\caption{DT Layer 0}
\label{fig:NNPSl1-xlmr}
\end{minipage}%
}%
\subfigure[Layer 8]{
\begin{minipage}[t]{0.32\linewidth}
\includegraphics[width=1\textwidth]{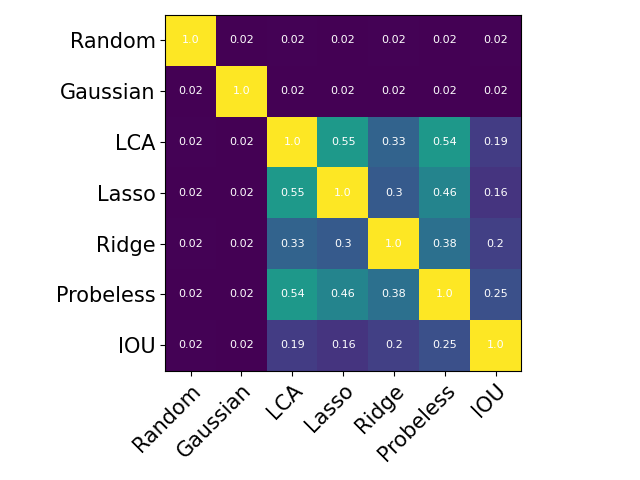}
%\caption{DT Layer 0}
\label{fig:NNPSl66-xlmr}
\end{minipage}%
}%
\subfigure[Layer 9]{
\begin{minipage}[t]{0.32\linewidth}
\includegraphics[width=1\textwidth]{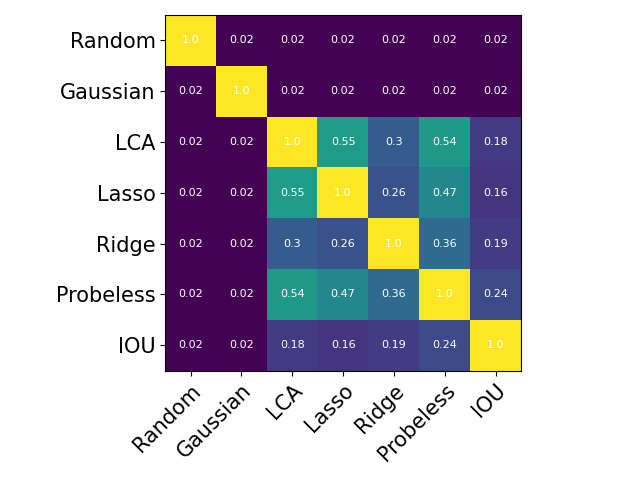}
%\caption{DT Layer 0}
\label{fig:NNPSl612-xlmr}
\end{minipage}%
}%
\\
% \subfigure[VBZ L0]{
% \begin{minipage}[t]{0.25\linewidth}
% \includegraphics[width=1\textwidth]{./figures/heatmap_layerVBZ_tag0_neuron_10.png}
% \end{minipage}
% }%
\subfigure[Layer 10]{
\begin{minipage}[t]{0.32\linewidth}
\includegraphics[width=1\textwidth]{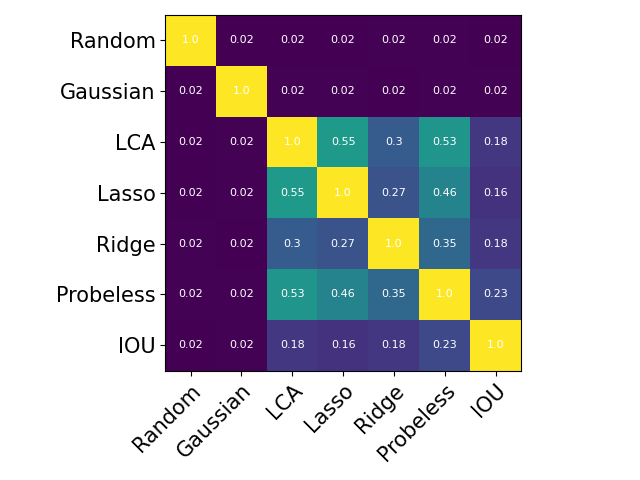}
%\caption{DT Layer 0}
\label{fig:VBZl6-xlmr}
\end{minipage}%
}%
\subfigure[Layer 11]{
\begin{minipage}[t]{0.32\linewidth}
\includegraphics[width=1\textwidth]{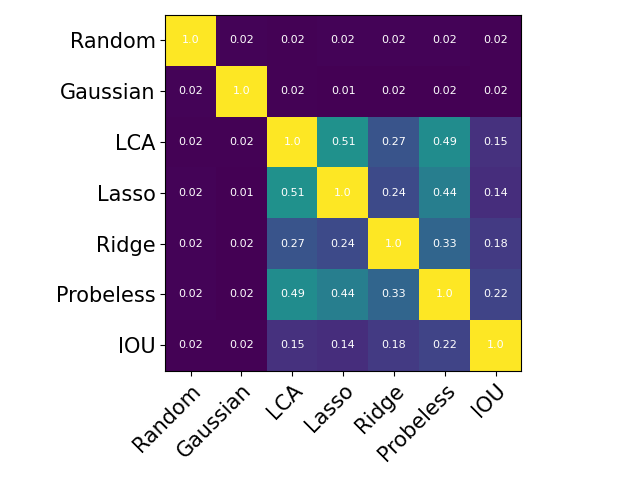}
%\caption{DT Layer 0}
\label{fig:VBZl6-xlmr-l11}
\end{minipage}%
}%
\subfigure[Layer 12]{
\begin{minipage}[t]{0.32\linewidth}
\includegraphics[width=1\textwidth]{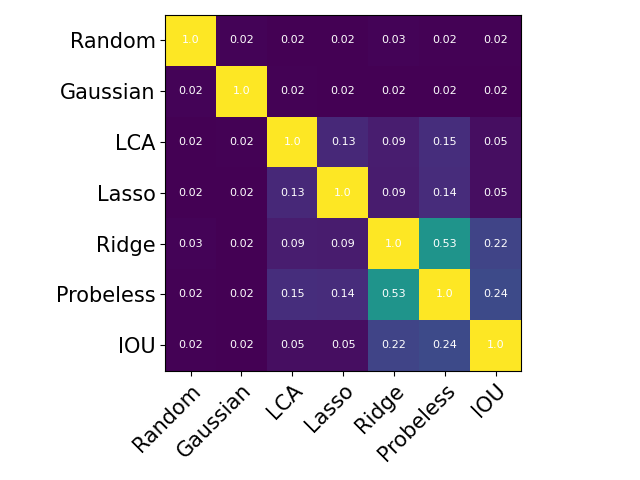}
%\caption{DT Layer 0}
\label{fig:VBZl62-xlmr}
\end{minipage}%
}%
\\
% \subfigure[VBZ L0]{
% \begin{minipage}[t]{0.25\linewidth}
% \includegraphics[width=1\textwidth]{./figures/heatmap_layerVBZ_tag0_neuron_10.png}
% \end{minipage}
% }%
% \subfigure[NN layer 1]{
% \begin{minipage}[t]{0.32\linewidth}
% \includegraphics[width=1\textwidth]{./figures/heatmap_layer1_NN_neuron_10.png}
% %\caption{DT Layer 0}
% \label{fig:NNl1}
% \end{minipage}%
% }%
% \subfigure[NN layer 6]{
% \begin{minipage}[t]{0.32\linewidth}
% \includegraphics[width=1\textwidth]{./figures/heatmap_layer6_NN_neuron_10.png}
% %\caption{DT Layer 0}
% \label{fig:NNl6}
% \end{minipage}%
% }%
% \subfigure[NN layer 12]{
% \begin{minipage}[t]{0.32\linewidth}
% \includegraphics[width=1\textwidth]{./figures/heatmap_layer12_NN_neuron_10.png}
% %\caption{DT Layer 0}
% \label{fig:NNl12}
% \end{minipage}%
% }
\centering
\caption{This is an extension of Figure.\ref{fig:top10_50_heatmap}. Comparing average overlap of top 10-50 neurons across methods for XLMR}
\label{app:fig:top10_heatmap_xlmr}
\end{figure*}
% Please add the following required packages to your document preamble:
% \usepackage{multirow}
% Please add the following required packages to your document preamble:
% \usepackage{multirow}

\begin{table*}[]
\caption{Comparison of NeuronVote score of Probeless and LCA under different number of consensus methods. In majority of the cases, Probeless outperformed LCA irrespective of the methods used for consensus. Moreover, in three cases where LCA is better than Probeless, their results are comparable.}
\centering
\footnotesize
\begin{tabular}{c|l|cc}
\hline
Num. Methods     & Consensus Methods & Probeless      & LCA            \\ \hline
1 & Gaussian                            & \textbf{0.110} & 0.096          \\
                   & IoU                                 & \textbf{0.274} & 0.216          \\
                   & Lasso                            & 0.449          & \textbf{0.516} \\
                   & Ridge                            & \textbf{0.270} & 0.240          \\ \hline
2 & Gaussian,IoU                        & \textbf{0.233} & 0.177          \\
                   & Gaussian,Ridge                   & \textbf{0.202} & 0.174          \\
                   & Gaussian,Lasso                   & 0.272          & \textbf{0.275} \\
                   & Lasso,IoU                        & \textbf{0.351} & 0.329          \\
                   & Lasso,Ridge                   & 0.404          & \textbf{0.410} \\
                   & Ridge,IoU                        & \textbf{0.292} & 0.243          \\ \hline
3 & Gaussian,Ridge,IoU                  & \textbf{0.265} & 0.229          \\
                   & Gaussian,Lasso,Ridge          & \textbf{0.337} & 0.327          \\
                   & Gaussian,Lasso,IoU               & \textbf{0.327} & 0.305          \\
                   & Lasso,Ridge,IoU               & \textbf{0.373} & 0.351          \\ \hline
4                  & Gaussian,Lasso,Ridge,IoU         & \textbf{0.368} & 0.342          \\ \hline
\end{tabular}
\end{table*}

\begin{table*}[]
\caption{Comparison of NeuronVote score of Probeless and Lasso under different number of consensus methods. In all cases, Probeless shows better results irrespective of the methods used for consensus.}
\centering
\footnotesize
\begin{tabular}{c|l|cc}
\hline
Num of Consensus Methods & Consensus Methods         & Probeless      & Lasso           \\ \hline
1       & Gaussian                  & \textbf{0.110} & 0.092          \\
                         & IoU                       & \textbf{0.274} & 0.195          \\
                         & LCA                       &  \textbf{0.524}       & 0.516 \\
                         & Ridge                  & \textbf{0.270} & 0.237          \\ \hline
2      & Gaussian,IoU              & \textbf{0.233} & 0.159          \\
                         & Gaussian,LCA              & \textbf{0.291} & 0.267          \\
                         & Gaussian,Ridge         & \textbf{0.202}          & 0.170 \\
                         & LCA,IoU                   & \textbf{0.357} & 0.300          \\
                         & LCA,Ridge              & \textbf{0.387}          & 0.329 \\
                         & Ridge,IoU              & \textbf{0.292} & 0.227          \\ \hline
3       & Gaussian,LCA,IoU          & \textbf{0.346} & 0.289          \\
                         & Gaussian,LCA,Ridge     & \textbf{0.356} & 0.321          \\
                         & Gaussian,Ridge,IoU     & \textbf{0.265} & 0.214          \\
                         & LCA,Ridge,IoU          & \textbf{0.424} & 0.396          \\ \hline
4                        & Gaussian,LCA,Ridge,IoU & \textbf{0.382} & 0.324          \\ \hline
\end{tabular}
\end{table*}
% Please add the following required packages to your document preamble:
% \usepackage{multirow}
\begin{table*}[]
\centering
\footnotesize
\caption{Comparison of NeuronVote score of Probeless and IoU under different number of consensus methods. In all cases, Probeless shows better results irrespective of the methods used for consensus.}
\begin{tabular}{c|l|cc}
\hline
Num of Consensus Methods & Consensus Methods           & Probeless      & IoU   \\ \hline
1      & Gaussian                    & \textbf{0.110} & 0.085 \\
                         & IoU                         & \textbf{0.449} & 0.195 \\
                         & Lasso-01                    & \textbf{0.524} & 0.216 \\
                         & Ridge-01                    & \textbf{0.270} & 0.160 \\ \hline
2       & Gaussian,IoU                & \textbf{0.272} & 0.155 \\
                         & Gaussian,Ridge-01           & \textbf{0.291} & 0.158 \\
                         & Gaussian,Lasso-01           & \textbf{0.202} & 0.136 \\
                         & Lasso-01,IoU                & \textbf{0.404} & 0.206 \\
                         & Lasso-01,Ridge-01           & \textbf{0.498} & 0.209 \\
                         & Ridge-01,IoU                & \textbf{0.424} & 0.217 \\ \hline
3       & Gaussian,Ridge,IoU          & \textbf{0.337} & 0.194 \\
                         & Gaussian,Lasso-01,Ridge-01  & \textbf{0.356} & 0.200 \\
                         & Gaussian,Lasso-01,IoU       & \textbf{0.475} & 0.224 \\
                         & Lasso-01,Ridge-01,IoU       & \textbf{0.434} & 0.206 \\ \hline
4                        & Gaussian,Lasso,Ridge-01,IoU & \textbf{0.426} & 0.221 \\ \hline
\end{tabular}
\end{table*}

\end{document}